\documentclass[opre,nonblindrev]{informs3_homepage}

\usepackage{url}

\usepackage[pagebackref=true]{hyperref}
\hypersetup{
	colorlinks=true,
	linkcolor=red,
	filecolor=blue,
	citecolor = blue,      
	urlcolor=black,
}

\usepackage{subcaption}
\usepackage{algorithm}
\usepackage{algorithmic}

\usepackage{smileInformsCompatible}
\usepackage{algorithm}
\usepackage{algorithmic}
\usepackage{color}
\usepackage{comment}
\usepackage{mathtools}
\usepackage{times}
\usepackage{authblk}
\usepackage{layout}
\usepackage{soul}

\usepackage{lmodern}

\def\##1\#{\begin{align}#1\end{align}}
\def\$#1\${\begin{align*}#1\end{align*}}

\newcommand{\lambdareg}{\lambda_{\text{reg}}}
\newcommand{\event}[1]{\II\{#1\}} 
\newcommand{\param}{\theta^\star}
\newcommand{\tparam}{\tilde{\theta}}
\newcommand{\hparam}{\hat{\theta}}
\newcommand{\halpha}{\hat{\alpha}}

\newcommand{\RLS}{RLS}
\newcommand{\PE}{PE}
\newcommand{\PVT}{PVT}

\usepackage{tikz}

\usetikzlibrary{shapes.geometric} 
\usetikzlibrary{shapes.geometric} 
\usetikzlibrary{decorations.pathreplacing, arrows.meta}

\usetikzlibrary{arrows.meta} 

\definecolor{pastelblue}{rgb}{0,0.3, 0.6}
\definecolor{axiscolor}{rgb}{0.7, 0.65, 0.6}
\definecolor{dashedlinecolor}{rgb}{0.5, 0.5, 0.5}

\def\##1\#{\begin{align}#1\end{align}}
\def\$#1\${\begin{align*}#1\end{align*}}

\newcommand{\POFUL}{POFUL}

\ECRepeatTheorems

\begin{document}

	
	
	
	\RUNTITLE{Geometry-Aware Bandit Algorithms}
	
	\TITLE{Geometry-Aware Approaches for Balancing Performance and Theoretical Guarantees in Linear Bandits}
	
	
	\ARTICLEAUTHORS{%
		\AUTHOR{Yuwei Luo}
		\AFF{Graduate School of Business, Stanford University, \EMAIL{yuweiluo@stanford.edu}}
		\AUTHOR{Mohsen Bayati}
		\AFF{
			Graduate School of Business, Stanford University, \EMAIL{bayati@stanford.edu}}
	} 

\ABSTRACT{
This paper is motivated by recent research in the $d$-dimensional stochastic linear bandit literature, which has revealed an unsettling discrepancy: algorithms like Thompson sampling and Greedy demonstrate promising empirical performance, yet this contrasts with their pessimistic theoretical regret bounds. The challenge arises from the fact that while these algorithms may perform poorly in certain problem instances, they generally excel in typical instances. To address this, we propose a new data-driven technique that tracks the geometric properties of the uncertainty ellipsoid around the main problem parameter. This methodology enables us to formulate a data-driven frequentist regret bound, which incorporates the geometric information, for a broad class of base algorithms, including Greedy, OFUL, and Thompson sampling. This result allows us to identify and ``course-correct" problem instances in which the base algorithms perform poorly. The course-corrected algorithms achieve the minimax optimal regret of order $\tilde{\cO}(d\sqrt{T})$ for a $T$-period decision-making scenario, effectively maintaining the desirable attributes of the base algorithms, including their empirical efficacy. We present simulation results to validate our findings using synthetic and real data.}

	
\maketitle

\section{Introduction}


Multi-armed bandits (MABs) provide a framework for studying the exploration-exploitation trade-off in sequential decision-making, where a decision-maker selects actions and observes uncertain rewards. This extends to contextual bandits with features or covariates, as shown in numerous applications \citep{langford2008theepoch,li2010contextual,tewari2017ads,zhou2020survey,villar2015multi,bastani2020online,cohen2020feature}. This paper focuses on a well-studied class of models that captures both MABs and contextual bandits as special cases while being amenable to theoretical analysis: the stochastic linear bandit (LB) problem. In this model, the problem parameter $\param$ represents an unknown vector in $\RR^d$, while the actions, also vectors in $\RR^d$, yield noisy rewards with a mean equal to the inner product of $\param$ and the chosen action. The objective of a policy is to maximize the cumulative reward based on the observed data up to the decision time. The policy's performance is measured by the cumulative regret, which quantifies the difference between the total expected rewards achieved by the policy and the maximum achievable expected reward. 

Achieving this objective necessitates striking a balance between exploration and exploitation. In the context of LB, this entails selecting actions that aid in estimating the true parameter $\param$ accurately while obtaining optimal rewards. Various algorithms based on the \emph{optimism principle} have been developed to address this challenge, wherein the optimal action is chosen based on the upper confidence bound (UCB) \citep{lai1985asymptotically, auer2002using, dani2008stochastic, rusmevichientong2010linearly}. Another popular strategy is Thompson sampling (TS), a Bayesian heuristic introduced by \cite{thompson1933likelihood} that employs randomization to select actions according to the posterior distribution of reward functions. Additionally, the Greedy policy that selects the myopically best action is shown to be effective in contextual bandits \citep{kannan2018smoothed,raghavan2018externalities,hao2020adaptive, bastani2021mostly}.

In the linear bandit setting, two regret types are considered. Bayesian regret treats parameter $\param$ as a random variable with a prior distribution, averaging regret over noise, algorithm randomness, and parameter randomness, measuring expected performance across parameter realizations. \cite{russo2014learning} and \cite{dong2018information} establish an $\tilde{\cO}(d \sqrt{T})$ upper bound for the  Bayesian regret of the Thompson Sampling (TS) heuristic, referred to as LinTS, matching the minimax optimal bound by \cite{dani2008stochastic}. Here, $\tilde{\cO}$ denotes asymptotic order up to polylogarithmic factors. Frequentist regret assumes fixed unknown parameter $\param$, averaging only over noise and algorithm randomness. The OFUL algorithm \citep{abbasi2011improved} achieves an optimal $\tilde{\cO}(d\sqrt{T})$ frequentist regret bound. However, TS-Freq, a frequentist LinTS variant with inflated posterior variance, only achieves $\tilde{\cO}(d\sqrt{dT})$ \citep{agrawal2013thompson,abeille2017linear}, suboptimal by factor $\sqrt{d}$. \citet{hamidi2020frequentist} confirms this inflation is necessary and LinTS's frequentist regret cannot be improved. The Greedy algorithm lacks theoretical guarantees for linear bandit problems \citep{lattimore2017end}, suggesting both LinTS and Greedy may perform suboptimally.

Despite the theoretical gaps, LinTS demonstrates strong empirical performance \citep{russo2018tutorial}, suggesting posterior distribution inflation may be unnecessary in most scenarios. Similarly, the Greedy algorithm performs well in typical cases \citep{bietti2021contextual}. While optimism-based algorithms are computationally expensive (generally NP-hard \citep{dani2008stochastic,russo2014learning,agrawal2019recent}), LinTS and Greedy maintain computational efficiency. This disparity between theoretical, computational, and empirical performance prompts two questions: Can we identify problematic instances for LinTS and Greedy in a data-driven way and apply "course-correction" to ensure competitive frequentist regret bounds? Can this be achieved while preserving their empirical performance and computational efficiency?
In this paper, we provide positive answers to both questions. Specifically, we make the following \emph{contributions}.

\noindent 1. We develop a real-time geometric analysis technique for the $d$-dimensional confidence ellipsoid surrounding $\param$. This method is crucial for maximizing the use of historical data, advancing beyond methods that capture only limited information from the confidence ellipsoid, such as a single numerical value. Consequently, this facilitates a more precise ``course-correction''.

\noindent 2.  We introduce a comprehensive family of algorithms, termed \emph{\POFUL{}} (encompassing OFUL, LinTS, TS-Freq, and Greedy as specific instances), and derive a general, data-driven frequentist regret bound for them. This bound is efficiently computable using data observed from previous decision epochs.

\noindent 3.   We introduce course-corrected variants of LinTS and Greedy that achieve minimax optimal frequentist regret. These adaptations maintain most of the desirable characteristics of the original algorithms.



\subsection{Other Related Literature}

Our work is closely related to three main research streams: methodological foundations of linear bandits, bandit algorithms utilizing spectral properties, and data-driven exploration techniques. While these works share some similarities with our approach, we highlight the key differences and the unique aspects of our methodology.

From a methodological perspective, our regret analysis builds upon the foundations laid by \cite{abbasi2011improved}, \cite{agrawal2013thompson}, and \cite{abeille2017linear}. However, a key distinguishing factor is that our approach does not rely on optimistic samples, which is a departure from previous methods. This means that the algorithms we study do not always choose actions that are expected to perform better than the true optimal action. By allowing non-optimistic samples, we avoid the need to inflate the posterior distribution, a requirement in the works of \cite{agrawal2013thompson} and \cite{abeille2017linear}.

Our use of spectral information in bandit algorithms bears some resemblance to the study of \emph{Spectral Bandits} \citep{valko2014spectral, kocak2014Spectral, kocak2020spectral, kocak2020Best}. These works represent arm rewards as smooth functions on a graph, leveraging low-rank structures to improve algorithmic performance and obtain regret guarantees independent of the number of actions. In contrast, our approach exploits the spectral properties of the action covariance matrix, which is distinct from graph spectral analysis. Moreover, our research tackles the broader context of stochastic linear bandits without assuming any low-rank structure.

Our work also shares conceptual similarities with research on exploration strategies \citep{russo2016information, kirschner2018heteroscedastic} and data-driven exploration reduction \citep{bastani2021mostly, pacchiano2020regret, hamidi2020frequentist, hamidi2020general}. However, our methodology and data utilization differ significantly. For instance, \cite{bastani2021mostly} focuses on the minimum eigenvalue of the covariance matrix, a single-parameter summary of the observed data, while \cite{hamidi2020general} uses information from one-dimensional reward confidence intervals. The work of \cite{hamidi2020frequentist} is more closely related to ours, as it employs spectral information to improve the performance of Thompson Sampling in linear bandits. They use a single summary statistic called the \emph{thinness coefficient} to decide whether to inflate the posterior. In contrast, our approach leverages the full geometric details of the $d$-dimensional confidence ellipsoid, harnessing richer geometric information.



\section{Setup and Preliminaries}\label{sec:setting}

\paragraph{Notations.} We use $\|\cdot\|$ to denote the Euclidean 2-norm. For a symmetric positive definite matrix $A$ and a vector $x$ of proper dimension, we let $\|x\|_{A}\coloneqq \sqrt{x^\top A x}$ be the weighted 2-norm (or the $A$-norm). We let $\dotp{\cdot}{\cdot}$ denote the inner product in Euclidean  space such that $\dotp{x}{y} = x^\top y$. For a $d$-dimensional matrix $V$,   we let $\lambda_1(V)\geq  \lambda_2(V) \geq  \dots \geq  \lambda_d(V)$ be the eigenvalues of $V$  arranged in decreasing order. 
We let $\cB_d $ denote the unit ball in $\RR^d$, and $\cS_{d-1} = \{x\in\RR^d:\|x\| =  1\}$ denote the unit hypersphere in  $\RR^d$. For an interger $N\geq 1$, we let $[N]$ denote the set $\{1,2,\dots,N\}$. We use the $\cO(\cdot	)$ notation to suppress problem-dependent constants, and the $\tilde{\cO}(\cdot)$ notation further suppresses polylog factors.

\paragraph{Problem formulation and assumptions.} We consider the stochastic linear bandit problem. Let $\theta^\star \in \RR^d$ be a fixed but unknown parameter. At each time $t\in[T]$, a policy $\pi$  selects action $x_t$ from a set of action  $\cX_t \subset \RR^d $ according to the past observations and receives a reward $r_t  = \dotp{x_t}{\theta^\star} + \varepsilon_t$, where $ \varepsilon_t$ is mean-zero noise with a distribution specified in Assumption \ref{assump:noise} below. We measure the performance of $\pi$ with the cumulative expected regret 
$
\cR(T) = \sum_{t= 1}^T\dotp{x^{\star}_t}{\theta^\star} -  \dotp{x_t}{\theta^\star}\,,
$
where $x^{\star}_t$ is the best action at time $t$, i.e., 
$
x^{\star}_t=\arg\max_{x \in \cX_t}\dotp{x}{\theta^\star}\,.
$
Let $\mathcal{F}_t$ be a $\sigma$-algebra generated by the history $(x_1, r_1, \ldots, x_t, r_t)$ and the prior knowledge, $\mathcal{F}_0$. Therefore, $\{\mathcal{F}_t\}_{t\ge 0}$ forms a filteration such that each $\mathcal{F}_t$ encodes all the information up to the end of period $t$.

We make the following assumptions that are standard in the relevant literature.

\begin{assumption}[Bounded parameter]\label{assump:bounded-parameter}
	The unknown parameter  $\param$ is bounded as  $\|\param\|\leq S$, where $S > 0$ is known.
\end{assumption}
\begin{assumption}[Bounded action sets]\label{assump:bounded-action-sets}
	The action sets $\{\cX_t\}$ are uniformly bounded and closed subsets of $\RR^d$, such that $\|x\|\leq X_t$ for all $x\in \cX_t$ and all $t\in[T]$, where $X_t$'s are known and $\sup_{t\geq 1}\left\{X_t\right\} < \infty $. 
\end{assumption}

\begin{assumption} [Subgaussian reward noise]\label{assump:noise}
	The noise sequence $\{\varepsilon_t\}_{t\ge 1}$ is conditionally mean-zero and $R$-subgaussian, where $R$ is known. Formally, 
	for all real valued $\lambda$, 
	$
	 \EE\left[e^{\lambda \varepsilon_t} | \cF_t\right] \leq \exp \left(\lambda^2 R^2 / 2\right)
	 $.
	 This condition implies that $\EE \left[\varepsilon_t  | \cF_t\right]=0$ for all $t\geq 1$.
\end{assumption}

\subsection{Regularized  Least Square and Confidence Ellipsoid}

In this subsection, we review the useful frequentist tools developed by \cite{abbasi2011improved} for estimating the unknown parameter $\theta^*$ in linear bandit (LB) problems.

Consider an arbitrary sequence of actions $(x_1,\dots,x_t)$ and their corresponding rewards $(r_1,\dots,r_t)$. In LB problems, the parameter $\theta^*$ is typically estimated using the regularized least squares (RLS) estimator. Let $\lambdareg$ be a fixed regularization parameter. The sample covariance matrix $V_t$ and the RLS estimate $\hat{\theta}_t$ are defined as follows:
\begin{equation}\label{eq:ridge}
	V_t=\lambdareg I_d +\sum_{s=1}^{t} x_s x_s^{\top}, \quad \hat{\theta}_t=V_t^{-1} \sum_{s=1}^{t} x_s r_{s}.
\end{equation}

The following proposition from \cite{abbasi2011improved} establishes that the RLS estimate $\hat{\theta}_t$ concentrates around the true parameter $\theta^*$ with high probability.

\begin{proposition}[Theorem 2 in \cite{abbasi2011improved}]\label{prop:beta}
	Let $\delta \in (0,1)$ be a fixed confidence level. Then, with probability at least $1-\delta$, it holds for all $x \in \mathbb{R}^d$ that
	\begin{align*}
		\|\widehat{\theta}_t-\theta^{\star}\|_{V_t} \leq \beta_{t,\delta, \lambdareg}^{RLS}, \quad 
		|\dotp{x}{\widehat{\theta}_t-\theta^{\star}}| \leq\|x\|_{V_t^{-1}} \beta_{t,\delta, \lambdareg}^{RLS}
	\end{align*}
	where the confidence bound $\beta_{t,\delta, \lambdareg}^{RLS}$ is defined as
	\begin{equation}\label{eq:beta_rls}
		\beta_{t,\delta, \lambdareg}^{RLS} = R \sqrt{2 \log (\lambdareg+t)^{d / 2} \lambdareg^{-d / 2}{\delta}^{-1}} + \sqrt{\lambdareg} S.
	\end{equation}
\end{proposition}

 Proposition~\ref{prop:beta} enables us to construct the following sequence of confidence ellipsoids.
\begin{definition}
	\label{def:rls}
	Fix $\delta\in (0,1)$. We define the RLS confidence ellipsoid as	
	\[
	\cE_{t,\delta, \lambdareg}^{RLS}= \{\theta \in \RR^d: \|\theta - \hparam_t\|_{V_t}\leq \beta_{t,\delta, \lambdareg}^{RLS} \}\,.
	\]
\end{definition}

The next proposition, known as the \emph{elliptical potential lemma}, plays a central role in bounding the regret. This proposition provides the key element in the work of \cite{abbasi2011improved}, showing that the cumulative prediction error incurred by the action sequence used to estimate $\theta^*$ is small.

\begin{proposition}[Lemma 11 in \cite{abbasi2011improved}] \label{prop:potential}
	If $\lambdareg >1$, for an arbitrary sequence $(x_1,\dots,x_t)$, it holds that
	$
	\sum_{s=1}^t\left\|x_s\right\|_{V_s^{-1}}^2 \leq 2 \log \frac{\operatorname{det}\left(V_{t+1}\right)}{\operatorname{det}(\lambdareg I)} \leq 2 d \log (1+\frac{t}{\lambdareg})\,.
	$

\end{proposition}

\section{\POFUL{} Algorithms}
\label{sec:POFUL}
In this section, we introduce \POFUL{} (Pivot OFUL), a generalized framework of OFUL. This framework enables a unified analysis of frequentist regret for common algorithms.

At a high level, \POFUL{} is designed to encompass the exploration mechanism of OFUL and LinTS. \POFUL{} takes as input a sequence of \emph{inflation} parameters $\{\iota_t\}_{t\in [T]}$,  feasible (randomized) \emph{pivots} $\{\tilde{\theta}_t\}_{t\in[T]}$ and \emph{optimism} parameters $\{\tau_t\}_{t\in [T]}$. The inflation parameters are used to construct confidence ellipsoids that contain $\{\tilde{\theta}_t\}_{t\in[T]}$ with high probability. This is formalized in the next definition.

\begin{definition} \label{def:tilde_beta}
Fix  $\delta \in (0,1)$ and  $\delta^\prime = \delta/2 T$. Given the inflation parameters $\{\iota_t\}_{t\in [T]}$, we call random variables $\{\tilde{\theta}_t\}_{t\in[T]}$  feasible pivots if for all $t\in[T]$,
$
	\PP[\tilde{\theta}_t \in \cE_{t,\delta^\prime, \lambdareg}^{\PVT} |\mathcal{F}_t] \geq 1 - \delta^\prime
$,
where we define the ``pivot ellipsoid'' as
$
 \cE_{t,\delta, \lambdareg}^{\PVT}  \coloneqq  \{\theta \in \RR^d: \|\theta - \hparam_t\|_{V_t}\leq \iota_t \beta^{RLS}_{t,\delta,\lambdareg} \}
$.
\end{definition}

At each time $t$, \POFUL{} chooses the action that maximizes the optimistic reward 
\#\label{eq:POFUL}
\tilde{x}_t = \argmax_{x\in\cX_t} \  \dotp{x}{\tilde{\theta}_{t}}  +   \tau_t  \|x\|_{V^{-1}_t}\beta_{t,\delta^\prime, \lambdareg}^{RLS}\,,
\#
as shown in a pseudocode representation in Algorithm~\ref{alg:POFUL} and illustrated in Figure~\ref{fig:POFUL_illustration}. 

Recall OFUL encourages exploration by introducing the uncertainty term $\tau_t  \|x\|_{V^{-1}_t}\beta_{t,\delta^\prime, \lambdareg}^{RLS}$ in the reward, while LinTS explores through random sampling within the confidence ellipsoid. We let \POFUL{} select an arbitrary pivot (which can be random) from $\cE_{t,\delta^\prime, \lambdareg}^{\PVT}$ and maximize the optimistic reward to encompass arbitrary exploration mechanisms within $\cE_{t,\delta^\prime, \lambdareg}^{\PVT}$. 

\begin{algorithm}
	\caption{\POFUL{} }\label{alg:POFUL}
	\begin{algorithmic}
		\REQUIRE $T$, $\delta$, $\lambdareg$,  $\{\iota_t\}_{t\in [T]}$, $\{\tau_t\}_{t\in [T]}$
		\STATE Initialize $ V_{0} \leftarrow \lambdareg I_d$, $\widehat{\theta}_{1} \leftarrow 0$,  $\delta^\prime \leftarrow \delta/2T$
		\FOR{t = 0, 1,  \dots, T }
		\STATE Sample a feasible pivot $\tilde{\theta}_{t} $ with respect to  $\iota_t$ according to Definition~\ref{def:tilde_beta}
		\STATE $\tilde{x}_{t} \leftarrow \argmax_{x\in\cX_t} \dotp{x}{\tilde{\theta}_{t}}  +   \tau_t  \|x\|_{V^{-1}_t}\beta_{t,\delta^\prime, \lambdareg}^{RLS}$
		\STATE Observe reward $r_{t}$
		\STATE $V_{t+1} \leftarrow V_{t}+\tilde{x}_{t}  \tilde{x}_{t}^{\top}$
		\STATE $\hat{\theta}_{t+1} \leftarrow V_{t+1}^{-1}\sum_{s=1}^{t} \tilde{x}_s r_{s}$.
		\ENDFOR
	\end{algorithmic}
\end{algorithm}

\begin{figure}[!b]
\centering
\begin{subfigure}[]{0.48\textwidth}
	\centering
 
	\begin{tikzpicture}[scale=0.5, every node/.append style={transform shape, font=\Large}]
		
		\def\originX{-8}
		\def\originY{0}
		
		\def\centerxhat{\originX+4}
		\def\centeryhat{\originY+4}
		
		\def\centerxtilde{\centerxhat-1}
		\def\centerytilde{\centeryhat+1.5}
		
		\def\radiusHatX{3} 
		\def\radiusHatY{1.5}
		\def\radiusTildeX{1.5} 
		\def\radiusTildeY{0.75}
		
		
		\filldraw [black] (\originX,\originY) circle (1pt) node[anchor=north east] {$\mathcal{O}$};
		\draw[->,axiscolor] (\originX,\originY+0) -- (\originX+8,\originY+0) node[below] {};
		\draw[->,axiscolor] (\originX+0,\originY) -- (\originX+0,\originY+7.5) node[left] {};
		
		\draw [rotate around = {135:(\centerxhat,\centeryhat)}] (\centerxhat,\centeryhat) ellipse (\radiusHatX cm and \radiusHatY cm);
		\filldraw [] (\centerxhat,\centeryhat) circle (0.5pt) node[inner sep = 0pt, anchor=south west, font=\Large] {$\hat{\theta}_t$}; 
		\node[] at (\centerxhat+3.2,\centeryhat-0.7) { \Large $\mathcal{E}_{t,\delta^\prime, \lambdareg}^{P V T}$};
		
		\draw [rotate around = {135:(\centerxtilde,\centerytilde)}, pastelblue] (\centerxtilde,\centerytilde) ellipse (\radiusTildeX cm and \radiusTildeY cm);
		\filldraw [] (\centerxtilde,\centerytilde) circle (0.5pt) node[inner sep = 0pt, anchor=south west, font=\Large] {$\tilde{\theta}_t$};

		\draw[line width=2pt] (\originX+9,\originY) -- (\originX+9,\originY+7.5);
		
		\node[text width=4cm, align=center, font=\Large] at (\originX+4, \originY+.5) {Parameter space: $\mathbb{R}^d$};

		
		\def\intervalAxisX{\centerxhat+8}
		\def\intervalAxisY{\centeryhat} 
		
		\node[text width=3cm, align=center, font=\Large] at (\originX+12, \originY+.5) {Reward space: $\mathbb{R}$};

		\draw[->,axiscolor] (\intervalAxisX,\originY+1) -- (\intervalAxisX,\originY+7.5) node[left] {};
		
		
		\pgfmathsetmacro{\radiusPVT}{(\radiusHatX)^2*sqrt(2)/2/sqrt((\radiusHatY)^2 + (\radiusHatX)^2) + (\radiusHatY)^2*sqrt(2)/2/sqrt((\radiusHatY)^2 + (\radiusHatX)^2)}
		
		\pgfmathsetmacro{\radiusOFUL}{(\radiusTildeX)^2*sqrt(2)/2/sqrt((\radiusTildeY)^2 + (\radiusTildeX)^2) + (\radiusTildeY)^2*sqrt(2)/2/sqrt((\radiusTildeY)^2 + (\radiusTildeX)^2)}

		\node[anchor=west, font=\Large] (POFUL) at (\intervalAxisX+2,\centerytilde+\radiusOFUL) {$\text{\POFUL}$};
		\draw[dashed] (POFUL) -- (\intervalAxisX+0,\centerytilde+\radiusOFUL); 
		
		\draw[decorate,decoration={brace, amplitude=10pt, aspect = 0.3, mirror}] (\intervalAxisX,\centeryhat-\radiusPVT) -- (\intervalAxisX,\centeryhat+\radiusPVT) node[pos = 0.3, xshift=20pt, anchor=west, font=\Large] { 2$\iota_t\|x\|_{V_t^{-1}} \beta_{t,\delta^\prime,\lambdareg}^{R L S}$};
		
		\filldraw [] (\intervalAxisX,\centerytilde) circle (1pt) node[inner sep = 2.5pt, anchor=  east, font=\Large] {$\langle\tilde{\theta}_t, x\rangle$};

		\draw[decorate,decoration={brace, amplitude=5pt, mirror,aspect=0.3},pastelblue] (\intervalAxisX,\centerytilde - \radiusOFUL) -- (\intervalAxisX, \centerytilde+\radiusOFUL) node[pos = 0.3, xshift= 10pt, anchor=west, font=\Large] { 2$\tau_t\|x\|_{V_t^{-1}} \beta_{t,\delta^\prime,\lambdareg}^{R L S}$};
		
		\filldraw [black] (\intervalAxisX,\intervalAxisY) circle (1pt) node[inner sep = 2.5pt, anchor= east, font=\Large] {$\langle\hat{\theta}_t, x\rangle$};
		
		\draw (\intervalAxisX-0.1,\centeryhat+\radiusPVT) -- (\intervalAxisX+0.1,\centeryhat+\radiusPVT) node[right]{};
		\draw (\intervalAxisX-0.1,\centeryhat-\radiusPVT) -- (\intervalAxisX+0.1,\centeryhat-\radiusPVT) node[right]{};  
		\draw (\intervalAxisX-0.1,\centerytilde+\radiusOFUL) -- (\intervalAxisX+0.1,\centerytilde+\radiusOFUL)[pastelblue] node [right,pastelblue]{};
		\draw (\intervalAxisX-0.1,\centerytilde-\radiusOFUL) -- (\intervalAxisX+0.1,\centerytilde-\radiusOFUL)[pastelblue] node[right,pastelblue]{};

		\draw[dashed, dashedlinecolor] (\originX,\centerytilde+\radiusOFUL) -- (\intervalAxisX+0,\centerytilde+\radiusOFUL); 
		\draw[dashed, dashedlinecolor] (\originX,\centerytilde-\radiusOFUL) -- (\intervalAxisX+0,\centerytilde-\radiusOFUL); 
		
		\draw[dashed,dashedlinecolor] (\originX,\centeryhat+\radiusPVT) -- (\intervalAxisX+0,\centeryhat+\radiusPVT); 
		\draw[dashed,dashedlinecolor] (\originX,\centeryhat-\radiusPVT) -- (\intervalAxisX+0,\centeryhat-\radiusPVT);     
		
	\end{tikzpicture}
	\caption{}
	\label{fig:POFUL_illustration}
\end{subfigure}
\hfill
\begin{subfigure}[]{0.48\textwidth}
	\centering
	\begin{tikzpicture}[scale=0.5, every node/.append style={transform shape, font=\Large}]
		
		\def\originX{-8}
		\def\originY{0}
		
		\def\centerxhat{\originX+4}
		\def\centeryhat{\originY+4}

		\def\centerxtilde{\centerxhat-1}
		\def\centerytilde{\centeryhat+1.5}
		
		\def\radiusHatX{3} 
		\def\radiusHatY{1.5}
		\def\radiusTildeX{1.5} 
		\def\radiusTildeY{0.75}

		
		\def\intervalAxisX{\centerxhat-2}
		\def\intervalAxisY{\centeryhat} 

		\pgfmathsetmacro{\radiusPVT}{(\radiusHatX)^2*sqrt(2)/2/sqrt((\radiusHatY)^2 + (\radiusHatX)^2) + (\radiusHatY)^2*sqrt(2)/2/sqrt((\radiusHatY)^2 + (\radiusHatX)^2)}
		
		\pgfmathsetmacro{\radiusOFUL}{(\radiusTildeX)^2*sqrt(2)/2/sqrt((\radiusTildeY)^2 + (\radiusTildeX)^2) + (\radiusTildeY)^2*sqrt(2)/2/sqrt((\radiusTildeY)^2 + (\radiusTildeX)^2)}

    \def\dodgeLength{.02}
    \def\PVTtickerlength{.09}
    \def\OFULtickerlength{.12}
    
		
		\pgfmathsetmacro{\radiusOFUL}{0.6}
		
		\def\intervalAxisXOFUL{\intervalAxisX}
		
		\draw[->,axiscolor] (\intervalAxisXOFUL,\originY) -- (\intervalAxisXOFUL,\originY+7.5) node[left] {};
		\node[anchor=west, font=\Large] (OFUL) at (\intervalAxisXOFUL+1,\centeryhat+\radiusOFUL) {$\text{OFUL}$};
		\draw[dashed] (OFUL) -- (\intervalAxisXOFUL+0,\centeryhat+\radiusOFUL);

		\filldraw [black] (\intervalAxisXOFUL,\intervalAxisY) circle (1pt) node[inner sep = 2.5pt, anchor= east, font=\Large] {$\langle\hat{\theta}_t, x\rangle$}; 
		\draw (\intervalAxisXOFUL-\PVTtickerlength,\centeryhat+\dodgeLength) -- (\intervalAxisXOFUL+\PVTtickerlength,\centeryhat+\dodgeLength) node[right]{};
		\draw (\intervalAxisXOFUL-\PVTtickerlength,\centeryhat-\dodgeLength) -- (\intervalAxisXOFUL+\PVTtickerlength,\centeryhat-\dodgeLength) node[right]{};   
		\draw (\intervalAxisXOFUL-\OFULtickerlength,\centeryhat+\radiusOFUL) -- (\intervalAxisXOFUL+\OFULtickerlength,\centeryhat+\radiusOFUL)[pastelblue] node[right,pastelblue]{};
		\draw (\intervalAxisXOFUL-\OFULtickerlength,\centeryhat-\radiusOFUL) -- (\intervalAxisXOFUL+\OFULtickerlength,\centeryhat-\radiusOFUL)[blue] node[right,pastelblue]{};	
		
        \def\intervalAxisXTS{\intervalAxisX + 4}
		
		\draw[->,axiscolor] (\intervalAxisXTS,\originY) -- (\intervalAxisXTS,\originY+7.5) node[left] {};

		\node[anchor=west, font=\Large] (TS) at (\intervalAxisXTS+1,\centerytilde) {$\text{TS}$};
		\draw[dashed] (TS) -- (\intervalAxisXTS+0,\centerytilde); 
		
		\filldraw [] (\intervalAxisXTS,\centerytilde) circle (1pt) node[inner sep = 2.5pt, anchor=  east, font=\Large] {$\langle\tilde{\theta}_t, x\rangle$};
		
		\filldraw [black] (\intervalAxisXTS,\intervalAxisY) circle (1pt) node[inner sep = 2.5pt, anchor= east, font=\Large] {$\langle\hat{\theta}_t, x\rangle$};

		\draw (\intervalAxisXTS-\PVTtickerlength,\centeryhat+\radiusPVT) -- (\intervalAxisXTS+\PVTtickerlength,\centeryhat+\radiusPVT) node[right]{};
		\draw (\intervalAxisXTS-\PVTtickerlength,\centeryhat-\radiusPVT) -- (\intervalAxisXTS+\PVTtickerlength,\centeryhat-\radiusPVT) node[right]{};  
		\draw (\intervalAxisXTS-\OFULtickerlength,\centerytilde+\dodgeLength) -- (\intervalAxisXTS+\OFULtickerlength,\centerytilde+\dodgeLength)[pastelblue] node [right,pastelblue]{};
		\draw (\intervalAxisXTS-\OFULtickerlength,\centerytilde-\dodgeLength) -- (\intervalAxisXTS+\OFULtickerlength,\centerytilde-\dodgeLength)[pastelblue] node[right,pastelblue]{};

		
		\def\intervalAxisXGreedy{\intervalAxisX+8}
		
		\draw[->,axiscolor] (\intervalAxisXGreedy,\originY) -- (\intervalAxisXGreedy,\originY+7.5) node[left] {};
		
		\node[anchor=west, font=\Large] (Greedy) at (\intervalAxisXGreedy+1,\intervalAxisY) {$\text{Greedy}$};
		\draw[dashed] (Greedy) -- (\intervalAxisXGreedy+0,\intervalAxisY); 
		
		\filldraw [black] (\intervalAxisXGreedy,\intervalAxisY) circle (1pt) node[inner sep = 2.5pt, anchor= east, font=\Large] {$\langle\tilde{\theta}_t, x\rangle$=$\langle\hat{\theta}_t, x\rangle$};
		
		\draw (\intervalAxisXGreedy-\OFULtickerlength,\centeryhat+2*\dodgeLength) -- (\intervalAxisXGreedy+\OFULtickerlength,\centeryhat+2*\dodgeLength)[pastelblue] node [right,pastelblue]{};
		\draw (\intervalAxisXGreedy-\OFULtickerlength,\centeryhat-2*\dodgeLength) -- (\intervalAxisXGreedy+\OFULtickerlength,\centeryhat-2*\dodgeLength)[pastelblue] node[right,pastelblue]{};

		\draw (\intervalAxisXGreedy-\PVTtickerlength,\centeryhat+\dodgeLength) -- (\intervalAxisXGreedy+\PVTtickerlength,\centeryhat+\dodgeLength) node[right]{};
		\draw (\intervalAxisXGreedy-\PVTtickerlength,\centeryhat-\dodgeLength) -- (\intervalAxisXGreedy+\PVTtickerlength,\centeryhat-\dodgeLength) node[right]{};

	\end{tikzpicture}
	\caption{}
	\label{fig:POFUL_special_cases}
 \end{subfigure}
 \caption{(a) POFUL algorithms illustration for general $\iota_t$ and  $\tau_t$. (b) Special cases: OFUL ($\iota_t = 0$, $\tau_t = 1$), TS ($\tau_t = 0$), and Greedy ($\iota_t = \tau_t = 0$).}
\end{figure}

We demonstrate that \POFUL{} encompasses OFUL, LinTS, TS-Freq, and Greedy as special cases, as illustrated in Figure \ref{fig:POFUL_special_cases}.
\begin{example}[OFUL]\label{ex:oful}
	For stochastic linear bandit problems, OFUL chooses actions by solving the optimization problem 
	$
	\max_{x\in\cX_t}  \   \dotp{x}{ \hat{\theta}_t}  + \|x\|_{V_t^{-1}}   \beta_{t,\delta^\prime, \lambdareg}^{RLS}
	$.
	Therefore, OFUL is a specially case of \POFUL{} where $\iota_t = 0$, $\tau_t = 1$ and $
	\tilde{\theta}_t = \hat{\theta}_t$, the center of the confidence ellipsoid, for all $t\in[T]$.
\end{example}

Before describing how TS can be derived as an instance of \POFUL{}, we introduce a definition.
\begin{definition}
Let $\delta \in (0,1)$. We define $\cD^{SA}(\delta)$ as a distribution satisfying 
	$	 
	\PP_{\eta \sim \cD^{SA}(\delta)}\left[\|\eta\| \leq 1\right] \geq 1 - \delta
	$.
\end{definition}

\begin{example}[TS]\label{ex:ts}	Linear Thompson Sampling (LinTS) algorithm is a generic randomized algorithm that samples from a distribution constructed from the RLS estimate at each step. At time $t$, LinTS samples as 
$
\tilde{\theta}_{t}  =\hat{\theta}_t +\iota^{TS}_t \beta_{t,\delta^\prime, \lambdareg}^{RLS}  V_t^{-1/2} \eta_t
$,
where $\delta^\prime = \delta/2T$,  $\iota^{TS}_t$ is inflation parameter controlling the scale of the sampling range, and $\eta_t$ is a random sample from a normalized sampling distribution $\cD^{SA}(\delta^\prime)$ that concentrates with high probability. LinTS is a special case of \POFUL{} where  $\iota_t = \iota^{TS}_t$, $\tau_t = 0$ and 
$
\tilde{\theta_t} = \hat{\theta}_t +\iota^{TS}_t \beta_{t,\delta^\prime, \lambdareg}^{RLS}  V_t^{-1/2} \eta_t
$.
Setting  $\iota_t = \tilde{\cO}(1)$ corresponds to the original LinTS algorithm, while  setting $\iota_t= \tilde{\cO}(\sqrt{d})$  corresponds to the frequentist variant of LinTS studied in \cite{agrawal2013thompson,abeille2017linear}, namely TS-Freq. This means TS-Freq inflates the posterior by a factor of order $\sqrt{d}$.
\end{example}

\begin{example}[Greedy]\label{ex:greedy}
Greedy is a special case of \POFUL{} with $\iota_t = \tau_t = 0$, $\tilde{\theta}_t = \hat{\theta}_t$,  $\forall t$.
\end{example}


\section{Frequentist Regret Analysis of \POFUL{}}
\label{sec:regret}
In this section, we present the frequentist regret analysis of POFUL algorithms. Proofs appear in Appendix~\ref{sec:other_proofs}. We first introduce high-probability concentration events.

\begin{definition}
	Fix $\delta \in(0,1)$ and $\delta^\prime = \delta/2T$. We define $\beta^{\PVT}_{t,\delta^\prime,\lambdareg}  \coloneqq \iota_t \beta^{\RLS}_{t,\delta^\prime,\lambdareg} $, $
\hat{\cA}_t \coloneqq \{ \forall s\leq t: \|\widehat{\theta}_t-\theta^{\star}\|_{V_t} \leq \beta_{t,\delta^\prime, \lambdareg}^{RLS} \},\quad
\tilde{\cA}_t  \coloneqq \{ \forall s\leq t: \|\tilde{\theta}_t-\hat{\theta}_t\|_{V_t} \leq \beta^{\PVT}_{t,\delta^\prime,\lambdareg} \}$, and $ \cA_t  \coloneqq \hat{\cA}_t \cap  \tilde{\cA}_t $.

\end{definition}
\begin{proposition}\label{prop:concentration}
		Under Assumptions~\ref{assump:bounded-parameter}, \ref{assump:bounded-action-sets} and \ref{assump:noise} , we have
$\prob{\cA_T} \geq 1 - \delta.
$
\end{proposition}

\subsection{An Data-Driven Regret Bound for \POFUL{}}

In the following, we condition on the event $\cA_T$  which holds with probability $1-\delta$. The following proposition bounds the instantaneous regret of \POFUL{}.
\begin{proposition}\label{prop:instantaneous}
Suppose ${\theta}^\star \in \cE_{t,\delta^\prime, \lambdareg}^{RLS} $ and $\tilde{\theta}_t \in \cE_{t,\delta^\prime, \lambdareg}^{\PVT}$, it holds that 
	\#\label{eq:upperbound}
\dotp{x_t^\star}{ \theta^{\star}}-\dotp{\tilde{x}_t}{\theta^{\star}}\leq (1+\iota_t-\tau_t)\|x^{\star}_t\|_{V^{-1}_t}\beta_{t,\delta^\prime,\lambdareg}^{RLS} +   (1+\iota_t+\tau_t)\|\tilde{x}_t\|_{V^{-1}_t}\beta_{t,\delta^\prime,\lambdareg}^{RLS}.
\#
\end{proposition}

Note that this upper bound is different from what's used in the optimism-based methods \citep{abbasi2011improved,agrawal2013thompson,abeille2017linear}, we reproduce their upper bound and discuss the relationship of our method and theirs in Appendix~\ref{sec:regret_decomp}.

On the right-hand side of \eqref{eq:upperbound}, since the oracle optimal action sequence  $\{x^{\star}_t\}_{t\in[T]}$ is unknown to the algorithm and is different from the action sequence  $\{\tilde{x}_t\}_{t\in[T]}$ played by \POFUL{}, one cannot apply Proposition~\ref{prop:potential} to bound the summation $\sum_{t=1}^T \|\tilde{x}_t\|_{V_t^{-1}}^2 $ and get an upperbound of the regret. To address this,  the key point to connect  $\{\tilde{x}_t\}_{t\in[T]}$ and $\{x^{\star}_t\}_{t\in[T]}$ with $V_t^{-1}$-norm. This motivates the following definition.

\begin{definition}
	For each $t\geq 1$, let $\tilde{x}_t$ and $x_t^\star$ respectively denote the action chosen by \POFUL{} and the optimal action.  We define the \emph{uncertainty ratio} at time $t$ as $\alpha_t \coloneqq \|x_t^\star\|_{V_t^{-1}}/\|\tilde{x}_t\|_{V_t^{-1}}    $. We also define the \emph{(instantaneous) regret proxy} at time $t$ as $\mu_t  \coloneqq  \alpha_t(1+\iota_t -\tau_t)+1+\iota_t+\tau_t$.

\end{definition}

Note that $ \dotp{x}{\widehat{\theta}_t-\theta^{\star}}\leq\|x\|_{V_t^{-1}} \beta_{t,\delta^\prime, \lambdareg}^{RLS}
$ holds with high probability, we have that $\|x\|_{V_t^{-1}} $ essentially determines the length of the confidence interval of the reward $\dotp{x}{\theta^{\star}}$.  Hence, $\alpha_t$ serves as the ratio of uncertainty degrees of the reward obtained by the optimal action $x^{\star}_t$ and the chosen action $\tilde{x}_t$.

The intuition behind the definition for $\mu_t$ is constructing a regret upper bound similar to that of OFUL.  Specifically,  Proposition~\ref{prop:instantaneous} indicates
$ \dotp{x_t^\star}{ \theta^{\star}}-\dotp{\tilde{x}_t}{\theta^{\star}} \leq \mu_t  \|\tilde{x}_t\|_{V_t^{-1}} \beta^{RLS}_{t,\delta^\prime, \lambdareg}
$,
and we can check that the instantaneous regret of OFUL satisfies 
$\dotp{x_t^\star}{ \theta^{\star}}-\dotp{\tilde{x}_t}{\theta^{\star}} \leq 2\|\tilde{x}_t\|_{V_t^{-1}}   \beta^{RLS}_{t,\delta^\prime, \lambdareg}$. In this sense, $\mu_t$ is a proxy of the instantaneous regret incurred by \POFUL{} at time $t$. Moreover, OFUL can be regarded as a \POFUL{} algorithm whose $\mu_t$ is fixed at 2, and we could extend the definition of $\alpha_t$ to OFUL by solving $\mu_t  =  \alpha_t(1+\iota_t -\tau_t)+1+\iota_t+\tau_t$ and set $\alpha_t =  1 $ for all $t\in [T]$ for OFUL (recall that in OFUL, $\iota_t = 0$ and $\tau_t = 1$ for all $t \in [T]$). 


The following Theorem connects $\{\mu_t\}_{t\in[T]}$ and $\cR(T)$. It provides an oracle but general frequentist regret upper bound for all \POFUL{} algorithms.

\begin{theorem}[Oracle frequentist regret bound for \POFUL{}] 
\label{thm:oracle_regret} Fix $\delta \in(0,1)$ and let $\delta^\prime = \delta/2T$. Under Assumptions~\ref{assump:bounded-parameter}, \ref{assump:bounded-action-sets} and \ref{assump:noise}, with probability $1-\delta$, 
\POFUL{} achieves a regret of
\#\label{eq:oracle_regret_wo_inflation}
\cR(T) \leq     \sqrt{2 d \left(\sum_{t=1}^T\mu_t^2\right)\log \left(1+\frac{T}{\lambdareg}\right) 
 }\beta_{T,\delta^\prime,\lambdareg}^{RLS}\,.
\#
\end{theorem}

\begin{remark} 
We call Theorem~\ref{thm:oracle_regret} an \emph{oracle} regret bound as $\{\mu_t\}_{t\in[T]}$ for general \POFUL{} depends on the unknown system parameter $\param$. In general, they cannot be calculated by the decision-maker. Nevertheless, note that $\iota_t$ and $\tau_t$ are chosen by the decision-maker, when we have computable upper bounds $\{\hat{\alpha}_t\}_{t\in[T]}$ for $\{\alpha_t\}_{t\in[T]}$, using $\mu_t^2\leq 2\alpha_t^2(1+\iota_t -\tau_t)^2+2(1+\iota_t+\tau_t)^2$, we could calculate upper bounds for $\{\mu_t\}_{t\in[T]}$ as well. Consequently, Theorem~\ref{thm:oracle_regret} instantly turns into a data-driven regret bound for \POFUL{} and could be utilized later for course correction, which will be the aim of the next section. When we additionally know that $1+\iota_t - \tau_t$ is non-negative, we would use the equality $\mu_t = \alpha_t(1+\iota_t -\tau_t)+1+\iota_t+\tau_t$ directly for the bound.
\end{remark}

\begin{remark}
In the Discussion section of \citet{abeille2017linear}, the authors introduce a concept similar to the reciprocal of our $\alpha_t$. They suggest that the necessity of proving LinTS samples are optimistic could be bypassed if for some $\alpha > 0$ LinTS samples $\tilde{\theta}_t$ such that $\|x^\star(\tparam_t)\|_{V_t^{-1}} \geq \alpha \|x^\star(\param_t)\|_{V_t^{-1}}$ with constant probability, where $x^\star(\tparam_t)$ and $x^\star(\param_t)$ represent the optimal actions corresponding to $\tparam_t$ and $\param_t$, respectively. They pose this as an open question regarding the possibility of relaxing the requirement of inflating the posterior. In the following section, we provide a positive answer to this question by studying the reciprocal of their $\alpha$ using geometric arguments. This investigation offers an explanation for the empirical success of LinTS without the need for posterior inflation.
\end{remark}

\section{A Data-Driven Approach}\label{sec:data-driven}

In this section, we present the main contribution of this work which provides a data-driven approach to calibrating \POFUL{}. Note that $\iota_t$ and $\tau_t$ are parameters of \POFUL{} that can be controlled by a decision-maker, the essential point is to find a computable, non-trivial upper bound $\hat{\alpha}_t$ for the uncertainty ratio $\alpha_t$, which turns into an upper bound $\hat{\mu}_t$ for the regret proxy $\mu_t$ that's deeply related to the frequentist regret of \POFUL{}.


We focus on scenarios where $\tau_t = 0$ for all $t\in [T]$. These include LinTS and variants such as TS-Freq, as well as Greedy - standard algorithms still lacking theoretical regret guarantees. Below, we construct upper bounds $\{\hat{\alpha}_t\}_{t\in[T]}$ for the continuous-action scenario. Bounds for discrete-action scenarios appear in Appendix \ref{sec:discrete_actions}.

\subsection{Continuous action space}
\label{sec:bound_sst_ratio}  


Our strategy capitalizes on geometric insights related to the properties of the confidence ellipsoids, providing upper bounds that can be computed efficiently.  For the sake of a better illustration, we consider 
 $\cX_t = \cS_{d-1}$ for all $t\in[T]$ for this scenario, where $\cS_{d-1} =\left\{x \in \mathbb{R}^d:\|x\|=1\right\}$ is the unit hypersphere in $\mathbb{R}^d$. This is a standard example of continuous action space, and is the same as the setting considered in \cite{abeille2017linear}. We remark that for this specific setting, the problem is still hard. This is because we don't have a closed-form solution for the set of potentially optimal actions.
 
 In this setting, the optimal action $x_t^\star(\theta )  \coloneqq \argmax_{x\in \cX_t} \dotp{x}{\theta}$ takes the form $ x^\star_t(\theta ) = \theta/\|\theta\|$. To upper bound $\alpha_t$, we consider respectively the smallest and largest value of $\|x^\star_t(\theta)\|_{V_t^{-1}}$ for $\theta$ in the confidence ellipsoids of $\theta$, namely, $\cE^{\RLS}_{t,\delta^\prime, \lambdareg}$ and $\cE^{\PVT}_{t,\delta^\prime, \lambdareg}$. Specifically, we have 
  \# \label{eq:alpha_hat_raw}
 \alpha_t \leq \frac{\sup _{\theta \in \cE_{t,\delta^\prime, \lambdareg}^{\RLS}}\|x^\star_t(\theta)\|_{V_t^{-1}}}{ \inf _{\theta \in \cE_{t,\delta^\prime, \lambdareg}^{\PVT}}\|x^\star_t(\theta)\|_{V_t^{-1}}}.
 \#

\begin{figure}[htp]
	\centering
	\begin{subfigure}[b]{0.45\textwidth}
	\includegraphics[width=0.9\textwidth]{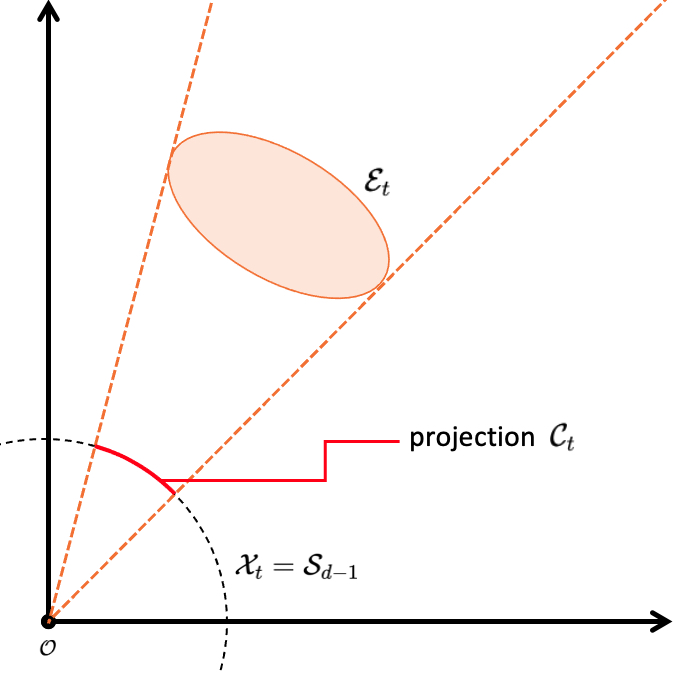} 
		\caption{}
	\end{subfigure}
\begin{subfigure}[b]{0.45\textwidth}
	\includegraphics[width=0.9\textwidth]{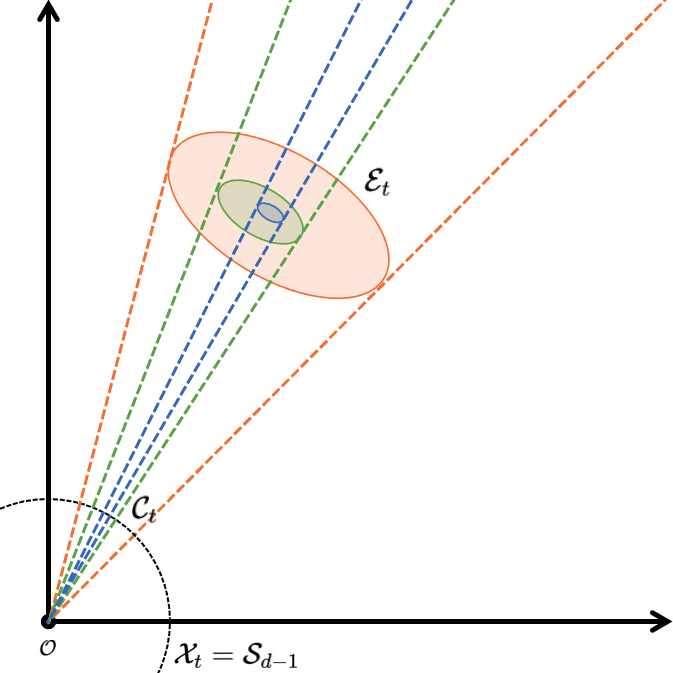} 
		\caption{}
	\end{subfigure}
	\caption{Illustration of potentially optimal actions set $\cC_t$ in $\RR^2$. (a): $\cC_t$ is $\cE_t$'s projection onto $\cS_{d-1}$. (b): As more data is collected, $\cE_t$ shrinks (colors show exploration levels). Potentially optimal actions point in similar directions, determining their $V_t$-norm. This suggests their $V_t$-norm range could be estimated geometrically. }
	\label{fig:ellipsoid1}
\end{figure}

As is illustrated in Figure~\ref{fig:ellipsoid1}, the set of potentially optimal actions $\cC_t$ is the projection of the confidence ellipsoid $\cE_t$  onto $\cS_{d-1}$. It's hard to get a closed-form expression for $\cC_t$, so we cannot directly calculate the range of $V_t^{-1}$-norm of actions in $\cC_t$. Nevertheless, when \POFUL{} has implemented sufficient exploration so that $\cE_t$ is small enough, $\cC_t$ concentrates accordingly to a small cap on $\cS_{d-1}$. Therefore, it is possible to estimate the range of the $V_t$-norm by employing geometric reasoning. Subsequently, this estimated range will be utilized to ascertain the range of the $V_t^{-1}$-norm.



The main theorem (proved in Section \ref{sec:proof_sketch_alpha}) derives an upper bound for $\alpha_t$ based on this idea.

\begin{theorem}\label{thm:alpha}

Suppose $\cX_t = \cS_{d-1}$ for all $t \in [T]$. Define $m_t = (\|\hat{\theta}_t\|_{V_t}^2  - (\beta^{RLS}_{t,\delta^\prime, \lambdareg})^2)/(\|\hat{\theta}_t\| +  \beta^{RLS}_{t,\delta^\prime, \lambdareg}/\lambda_d(V_t) )^2$, $ M_t = \|\hparam_t\|_{V_t^2}^2/(\|\hat{\theta}_t\|_{V_t}^2  - (\beta^{\PVT}_{t,\delta^\prime, \lambdareg})^2)$. Let  $k\in[d]$ be the integer that satisfies $\lambda_{k}(V) \leq M_t \leq \lambda_{k+1}(V) $.
 Define $\beta^{\PVT}_{t,\delta^\prime, \lambdareg} \coloneqq \iota_t \beta^{RLS}_{t,\delta^\prime, \lambdareg}$ and
					\$
	\Phi_t &=   \begin{cases}
				( \lambda_{1}^{-1}(V_t) + \lambda_{d}^{-1}(V_t)   - m_t\lambda_{1}^{-1}(V_t)\lambda_{d}^{-1} (V_t))^{\frac{1}{2}}, &\ \ \ \ \textit{if}\  \|\hat{\theta}_t\|_{V_t} \geq \beta^{RLS}_{t,\delta^\prime, \lambdareg}	\\
				\lambda_d^{-\frac{1}{2}}(V_t), & \ \ \ \ \textit{if}\  \|\hat{\theta}_t\|_{V_t} <  \beta^{RLS}_{t,\delta^\prime, \lambdareg} \end{cases},\\
	\Psi_t &=   \begin{cases}
				(\lambda_{k}^{-1}(V_t) +  \lambda_{k+1}^{-1}(V_t)   - M_t\lambda_{k}^{-1}(V_t)\lambda_{k+1}^{-1} (V_t))^{\frac{1}{2}}, &\textit{if}\  \|\hat{\theta}_t\|_{V_t} \geq \beta^{\PVT}_{t,\delta^\prime, \lambdareg}	\\
				\lambda_1^{-\frac{1}{2}}(V_t), & \textit{if}\  \|\hat{\theta}_t\|_{V_t} <  \beta^{\PVT}_{t,\delta^\prime, \lambdareg}\end{cases}.
				\$			

	Then for all $t \in [T]$, conditioned on $\hat{\cA}_t \cap \tilde{\cA}_t$, it holds for all $s \leq t$ that $\alpha_s\leq \halpha_s\coloneqq \Phi_s/\Psi_s$.


\end{theorem}
To better understand what Theorem~\ref{thm:alpha} implies, we discuss some special cases in Appendix~\ref{sec:Cases} and provide empirical validations for them.





\section{A Meta-Algorithm for Course-Correction}
\label{sec:meta}

This section demonstrates how the data-driven regret bound can enhance standard bandit algorithms. We propose a meta-algorithm that creates course-corrected variants of base algorithms, achieving minimax-optimal frequentist regret guarantees while preserving most original characteristics, including computational efficiency and typically low regret.

We take LinTS as an example of the base algorithm, and propose the algorithm Linear Thompson Sampling with Maximum Regret (Proxy) (TS-MR). The idea is to measure the performance of LinTS using $\hat{\mu}_t$ and avoid bad LinTS actions by switching to OFUL actions. Specifically, at each time $t$, TS-MR calculates the upper bound $\hat{\mu}_t$ and compares it with a preset threshold $\mu$. If  $\hat{\mu}_t > \mu$, LinTS might be problematic and  TS-MR takes an OFUL action to ensure a low instantaneous regret; if $\hat{\mu}_t\leq \mu$, TS-MR  takes the LinTS action. We remark that setting $\iota_t = 0$ for all $t\in [T]$ yields the corresponding Greedy-MR algorithm. The pseudocode is presented in Algorithm~\ref{alg:TS-MR} in Appendix~\ref{sec:TS-MR}. 

\begin{remark}
Computing $\hat{\mu}_t$ primarily requires SVD decomposition of the sample covariance matrix $V_t$. Since $V_t=\lambda_{\text{reg}} I_d+\sum_{s=1}^t x_s x_s^{\top}$ is updated via rank-one matrices, its SVD can be efficiently updated \citep{gandhi2017updating}, preventing computational bottlenecks.
\end{remark}

By design, course-corrected algorithms ensure that $\mu_t \leq \max\{\mu,2\}$ for all $t \in [T]$. Substituting this upper bound into Theorem~\ref{thm:oracle_regret} establishes that these algorithms achieve optimal frequentist regret, up to a constant factor.
\begin{corollary} TS-MR and Greedy-MR achieve a frequentist regret of
$\tilde{\cO}( \max\{\mu,2\} d \sqrt{T})$.
\end{corollary}


In high-risk settings where LinTS and Greedy may fail, a small $\mu$ ensures TS-MR and Greedy-MR select more OFUL actions, promoting sufficient exploration. Conversely, in low-risk settings where the original algorithms perform well, a large $\mu$ favors TS and greedy actions, minimizing unnecessary exploration and reducing computational cost. In Appendix~\ref{sec:tune_mu}, we show how $\mu$ impacts the fraction of OFUL actions in TS-MR and Greedy-MR and their performance. Results indicate that course-corrected algorithms maintain robustness across a range of moderate $\mu$ values, suggesting that the precise selection of $\mu$ is unlikely to present a significant practical concern.



\section{Simulations}
\label{sec:simulation}

We aim to compare TS-MR, Greedy-MR, and key baseline algorithms, via simulation.

\subsection{Synthetic datasets}
\label{sec:simulation_synthetic}

\begin{figure}[!htp]
	\centering
	\begin{subfigure}[b]{0.32\textwidth}
		\includegraphics[width=\textwidth]{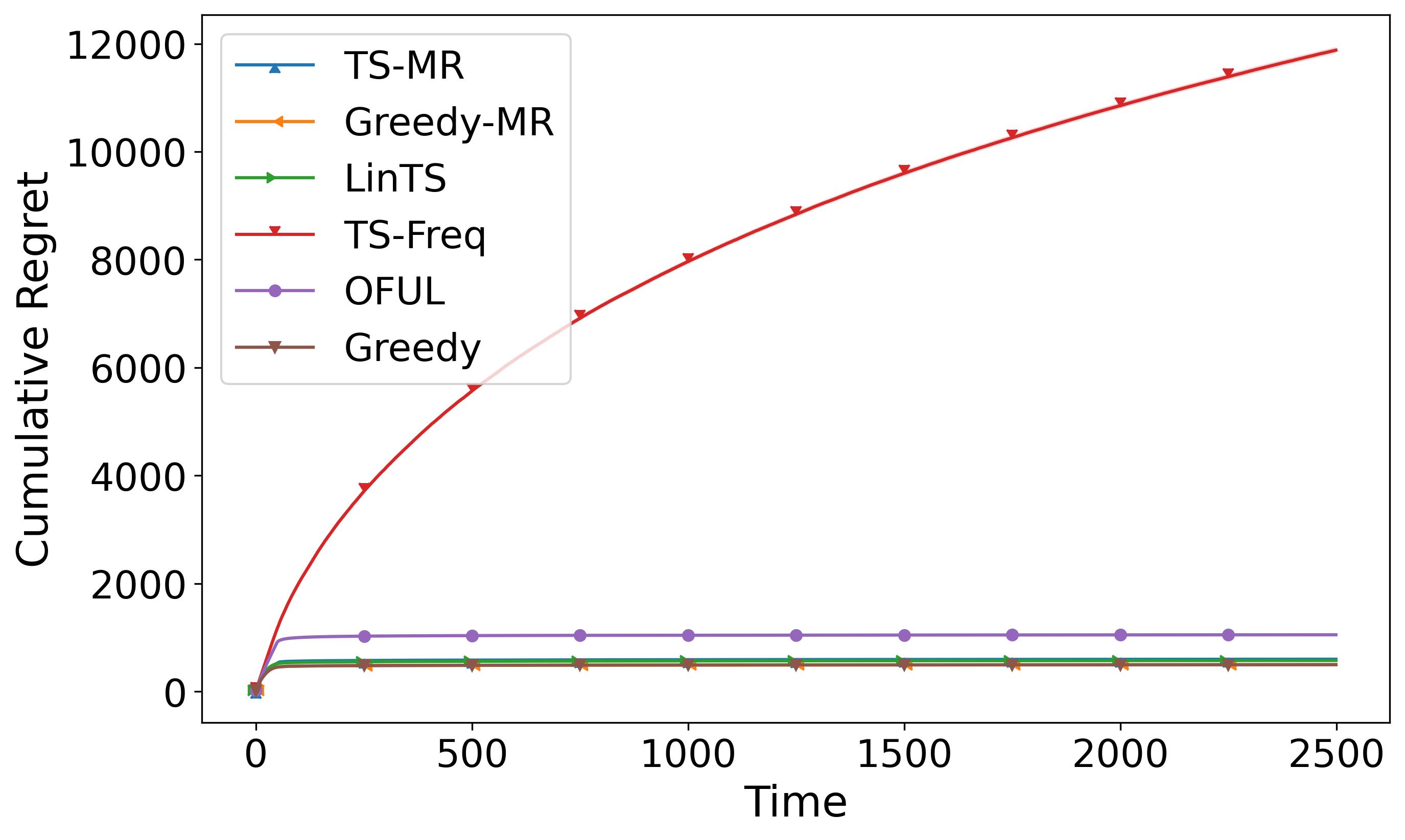}
				\caption{example 1}
				\label{fig:ex1}
\end{subfigure}
	\begin{subfigure}[b]{0.32\textwidth}
				\includegraphics[width=\textwidth]{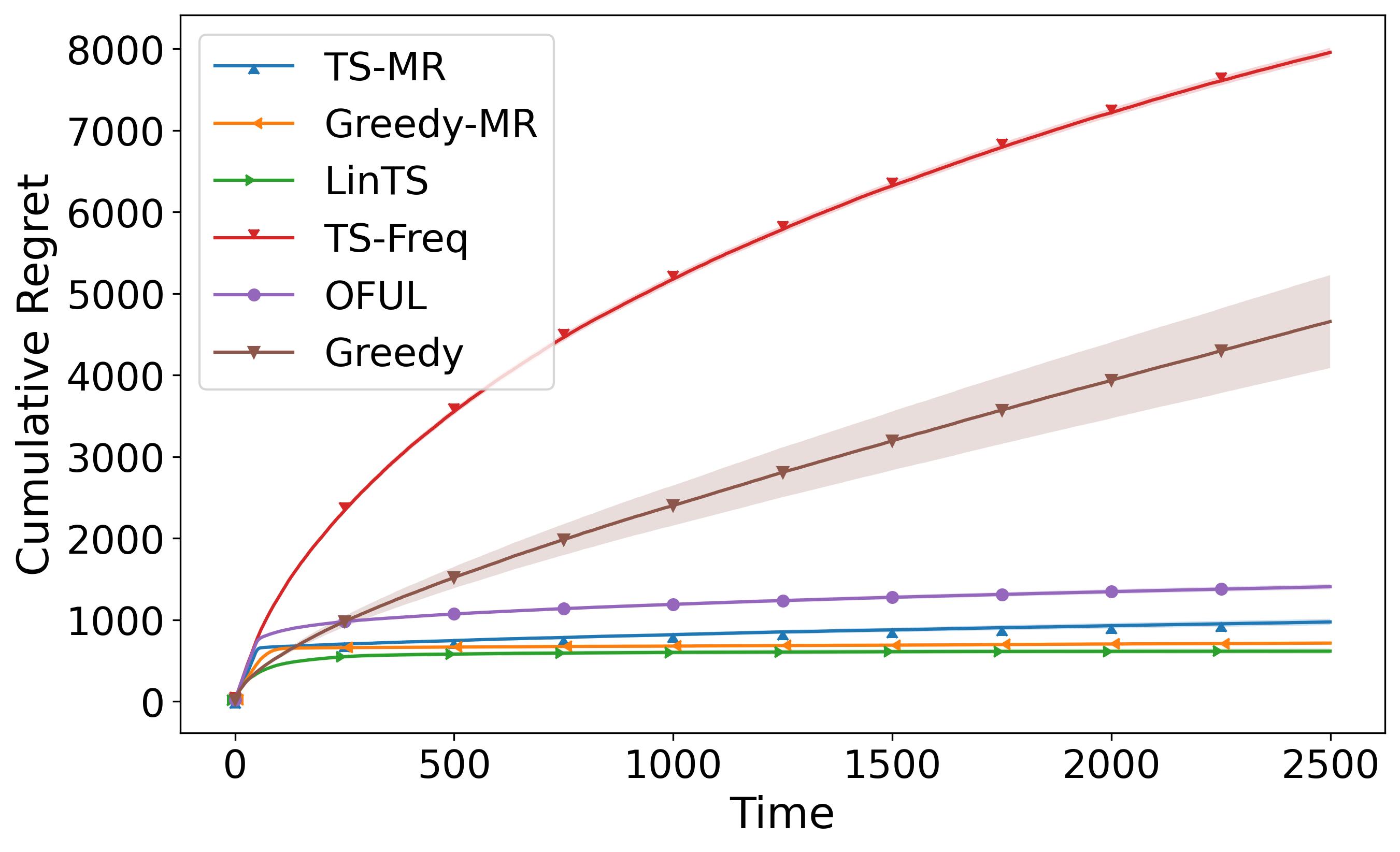}
						\caption{example 2}
						\label{fig:ex2}
\end{subfigure}
	\begin{subfigure}[b]{0.32\textwidth}
		\includegraphics[width=\textwidth]{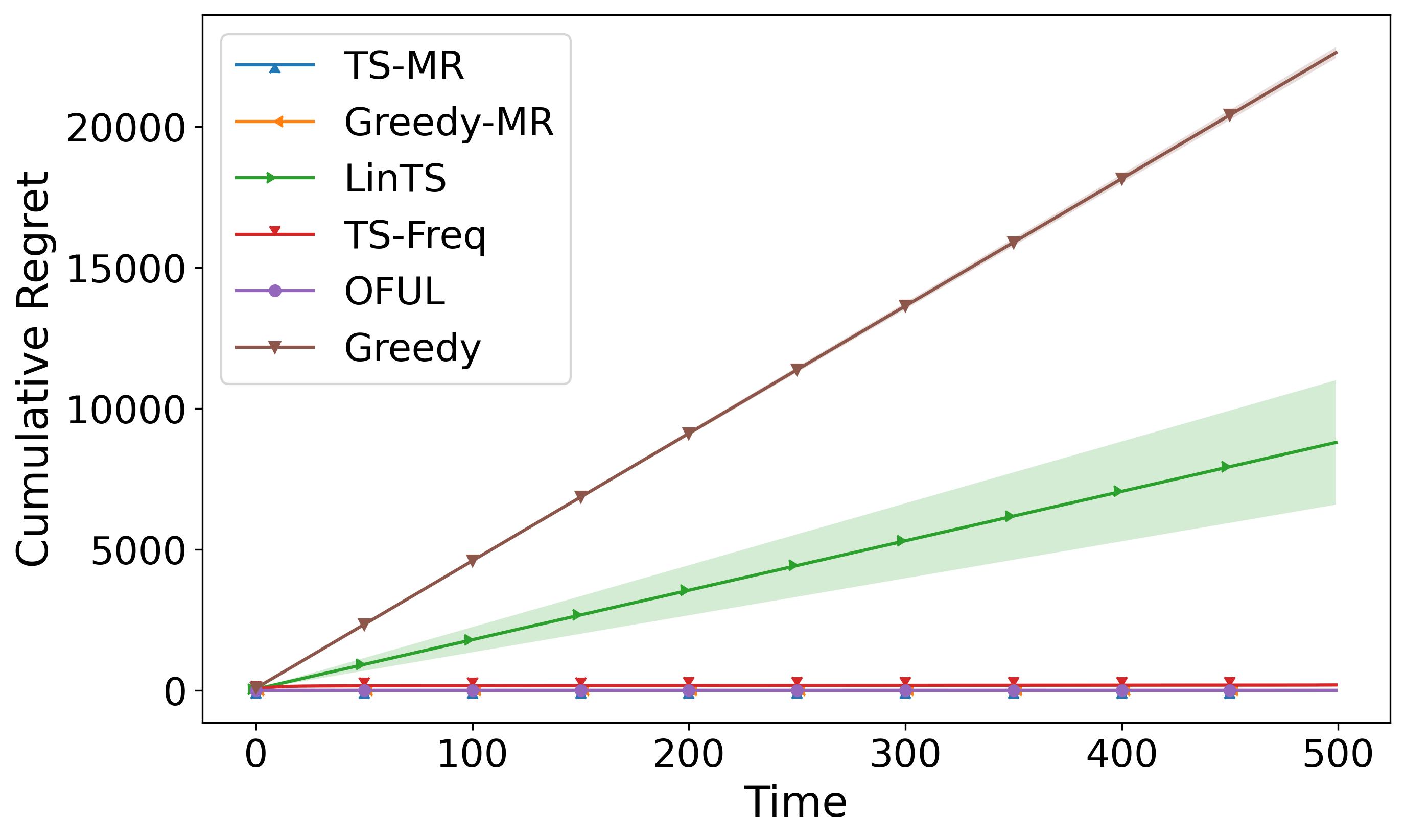}
				\caption{example 3}
				\label{fig:ex3}
\end{subfigure}\\
	\begin{subfigure}[b]{0.32\textwidth}
		\includegraphics[width=\textwidth]{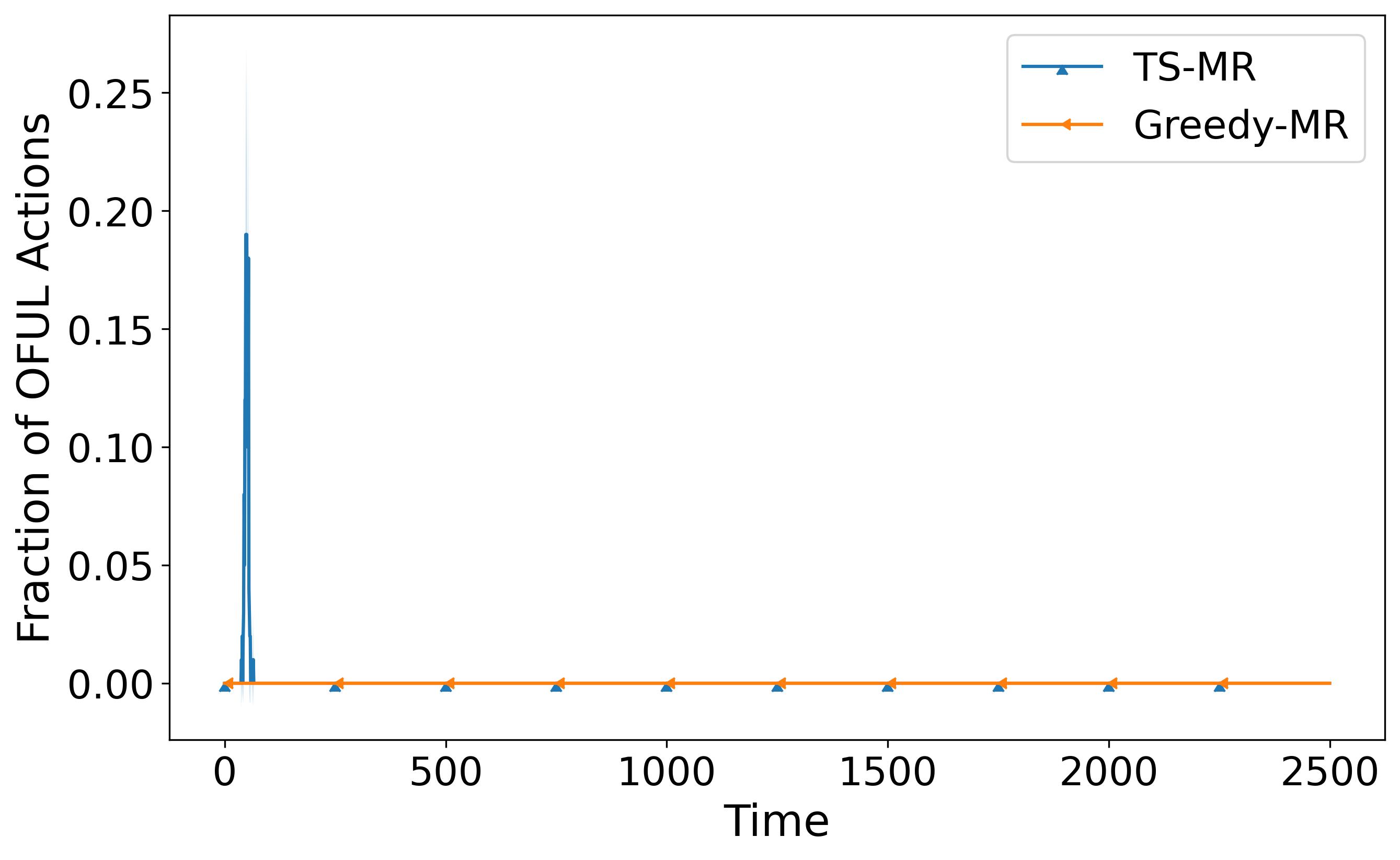} 
	\caption{example 1}
		\label{fig:oful_fraction_ex1}
\end{subfigure}
	\begin{subfigure}[b]{0.32\textwidth}
	\includegraphics[width=\textwidth]{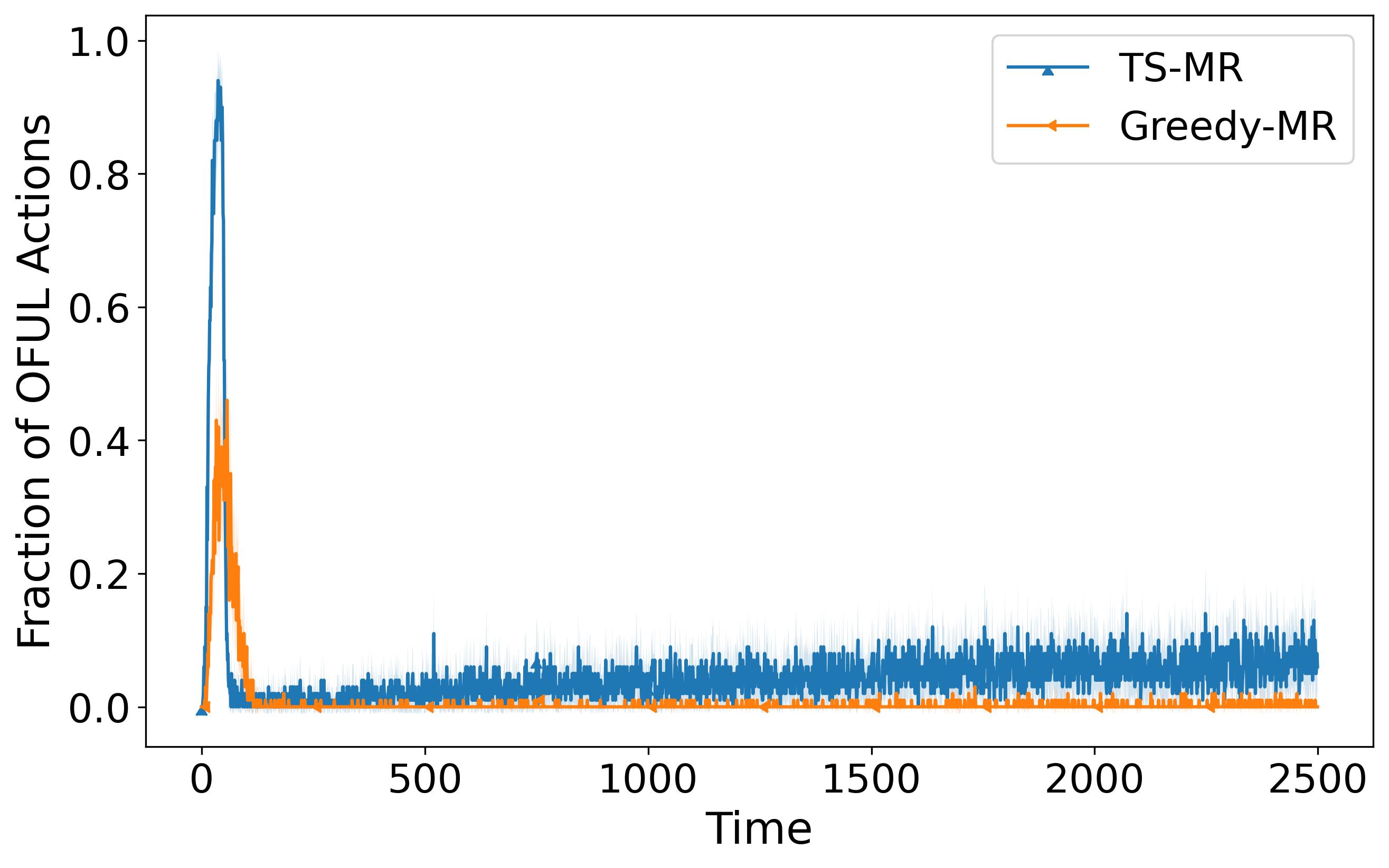} 
	\caption{example 2}
		\label{fig:oful_fraction_ex2}
	\end{subfigure}
 	\begin{subfigure}[b]{0.32\textwidth}
	\includegraphics[width=\textwidth]{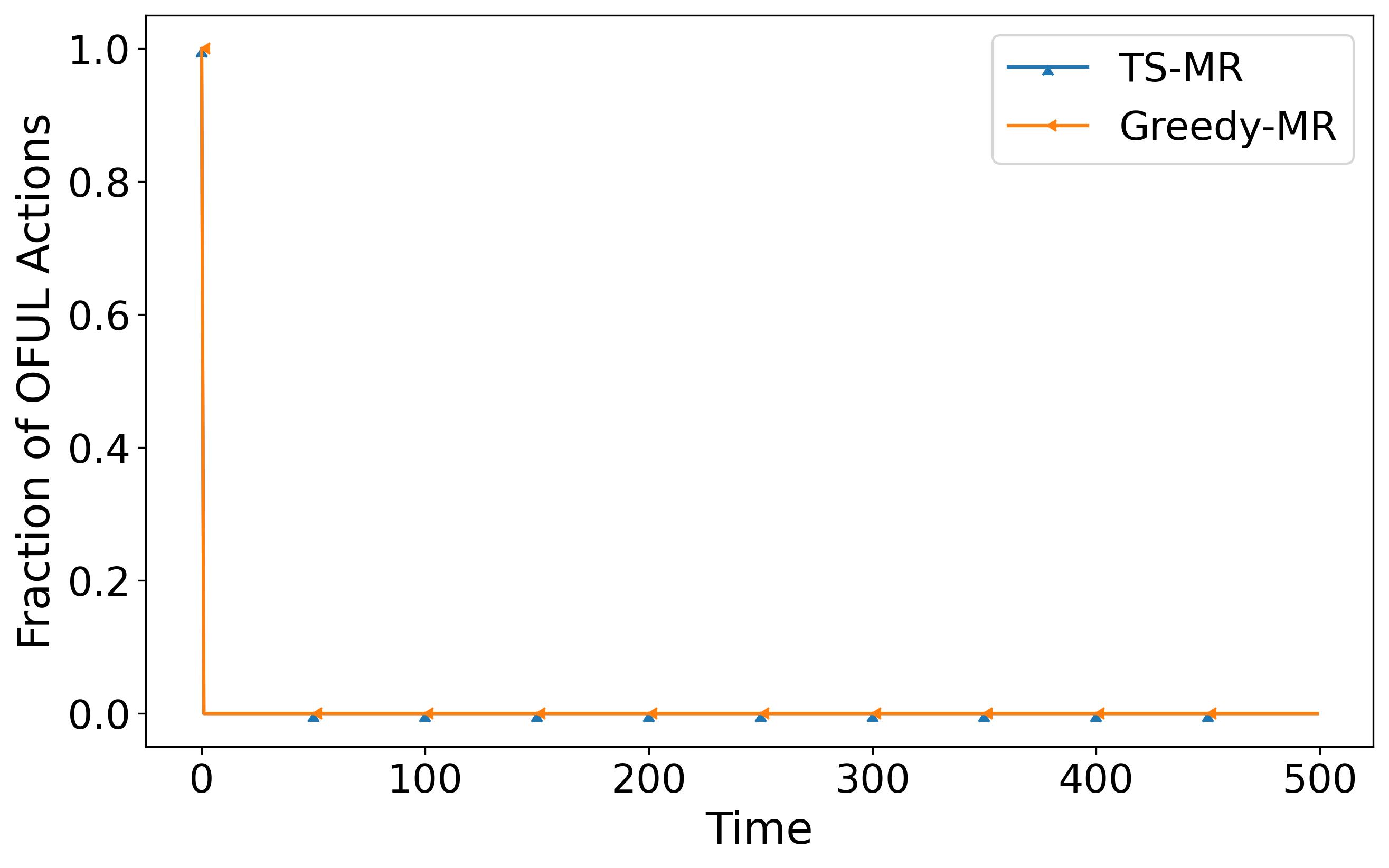} 
	\caption{example 3}
		\label{fig:oful_fraction_ex3}
	\end{subfigure}
	\caption{Simulation results on synthetic data. (a) - (c): Cumulative regret of TS-MR and Greedy-MR versus baseline algorithms. Shaded regions show $\pm 2$ SE of mean regret. (d) - (f): Fraction of OFUL actions in TS-MR and Greedy-MR.}
	\label{fig:synthetic}
\end{figure}

We conduct simulations on three representative synthetic examples.  We average simulation results over 100 independent runs for each of the examples. The results are shown in Figure~\ref{fig:synthetic}.




\paragraph{Example 1. Stochastic linear bandit with uniformly and independently distributed actions.} We fix $d = 50$, and sample $\param\sim\text{Unif}(\{\theta \in \mathbb{R}^d|\|\theta\|= 10\}) $ on a sphere with fixed norm. At each time $t$, we generate 100 i.i.d. random actions sampled from $\text{Unif}(\cS_{d-1})$ to form $\cX_t$. This is a basic example of standard stochastic linear bandit problems without any extra structure. We set the threshold  $\mu = 8$ for TS-MR and Greedy-MR. TS-Freq shows pessimistic regret due to the inflation of the posterior, while other algorithms in general perform well.

\paragraph{Example 2. Contextual bandits embedded in the linear bandit problem  \citep{abbasi2013online}.} We fix $d = 50$, and sample $\param\sim\text{Unif}(\{\theta \in \mathbb{R}^d|\|\theta\|= 70\}) $. At each time $t$, we first  generate a random vector $u_t \sim \cN(0, I_5)$ and let $x_{t,i}\in \RR^{50}$ be the vector whose $i$-th block of size 5 is a  copy of $u_t$, and other components are 0. Then $X_t = \{x_{t,i}\}_{i\in [10] }$ is an action set of size 10, sharing the same feature $u_t$ in different blocks. This problem is equivalent to a 10-armed contextual bandit. We set $\mu = 12$ for TS-MR and Greedy-MR.  In this setting, Greedy performs suboptimally due to a lack of exploration for some arms. Nevertheless, Greedy-MR outperforms both Greedy and OFUL by adaptively choosing OFUL actions only when it detects large regret proxy $\hat{\mu}_t$.

\paragraph{Example 3. Prior mean mismatch \citep{hamidi2020frequentist}.} This is an example in which LinTS is shown to incur linear Bayesian regret. We sample $\theta^\star \sim \cN(m \mathbf{1}_{3d}, I_{3d})$ and fix the action set $\cX_t = \{0, x_a, x_b\}$ for all $t\in[T]$, where $x_a = - \sum_{i=1}^{d} e_i/\sqrt{3d},  \quad x_b = \sum_{i=11}^{3d} e_i\sqrt{3d} - \sum_{i=1}^{d}e_i /\sqrt{3d}$.
It is shown in \cite{hamidi2020frequentist} that, when LinTS takes a wrong prior mean as input,  it has a large probability to choose  $\tilde{x}_2 = 0$, conditioned on $\tilde{x}_1 = x_a$. Note that choosing the zero action brings no information update to LinTS, it suffers a linear Bayesian regret when trying to escape from the zero action. We let $m = 10$ and set $d = 10$, so the problem is a $30$-dimensional linear bandit. We set $\mu = 12$ for TS-MR and Greedy-MR. We see both LinTS and Greedy incur linear regrets as expected, while TS-MR and Greedy-MR, switch to OFUL adaptively to tackle this hard problem and achieve sublinear regret.

\subsection{Real-world datasets}
\label{sec:simulation_real_world}
We explore the performance of standard \POFUL{} algorithms and the proposed TS-MR and Greedy-MR algorithms on real-world datasets. We use three classification datasets from \href{https://www.openml.org/}{\textsc{OpenML}}: \href{https://www.openml.org/d/1466}{Cardiotocography}, \href{https://www.openml.org/d/375}{JapaneseVowels}, and \href{https://www.openml.org/d/36}{Segment}, representing healthcare, pattern recognition, and computer vision domains. Following \citet{bietti2021contextual, bastani2021mostly}, we convert these classification tasks to contextual bandit problems and embed them into linear bandit problems as in Example 2, Section \ref{sec:simulation_synthetic}. Each class becomes an action where the decision-maker receives a binary reward (1 for correct classification, 0 otherwise) plus Gaussian noise.

\begin{figure}[!htp]
	\centering
	\begin{subfigure}[b]{0.32\textwidth}
		\includegraphics[width=\textwidth]{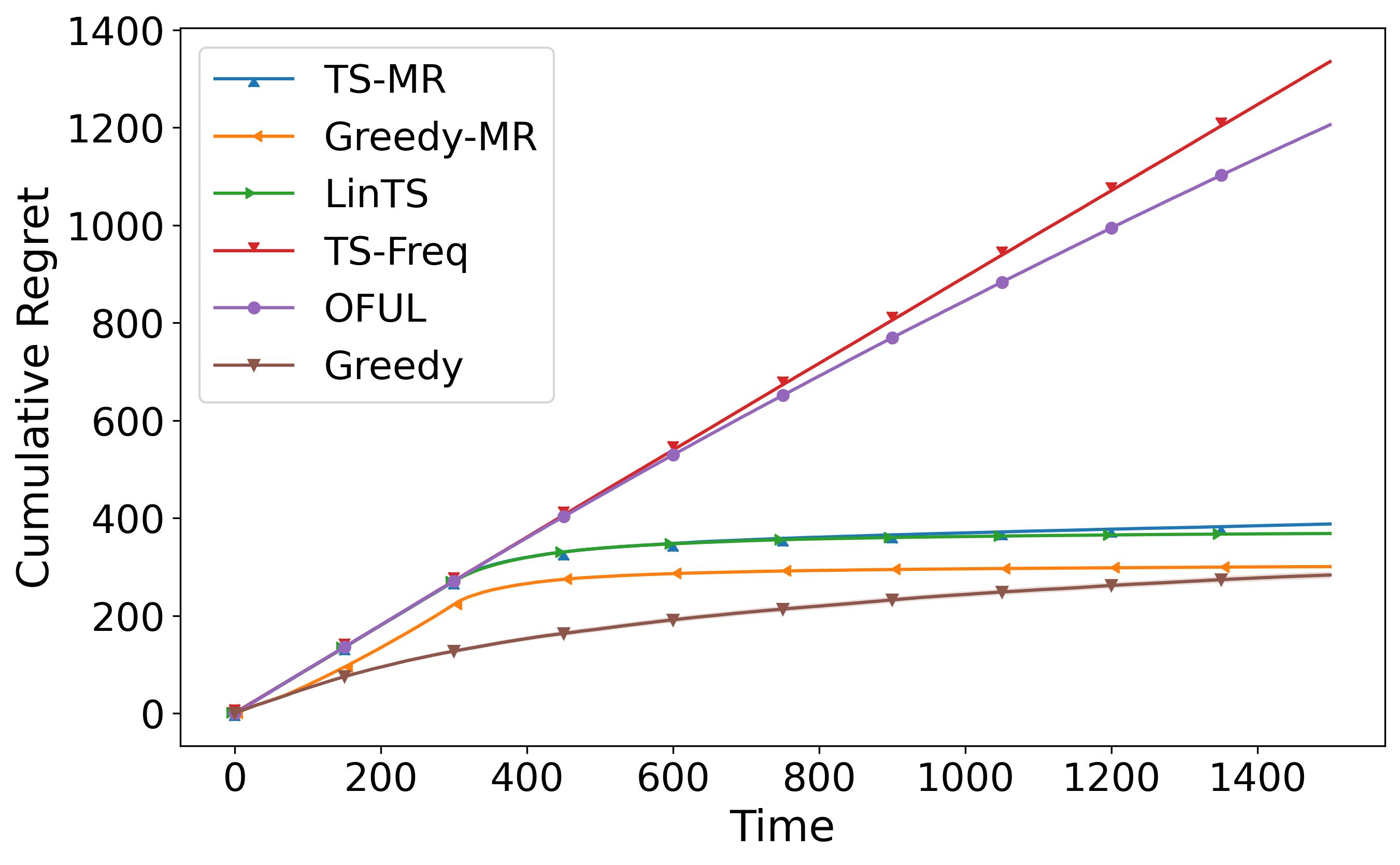} 
	\caption{Cardiotocography dataset}
		\label{fig:Cardiotocography}
\end{subfigure}
	\begin{subfigure}[b]{0.32\textwidth}
	\includegraphics[width=\textwidth]{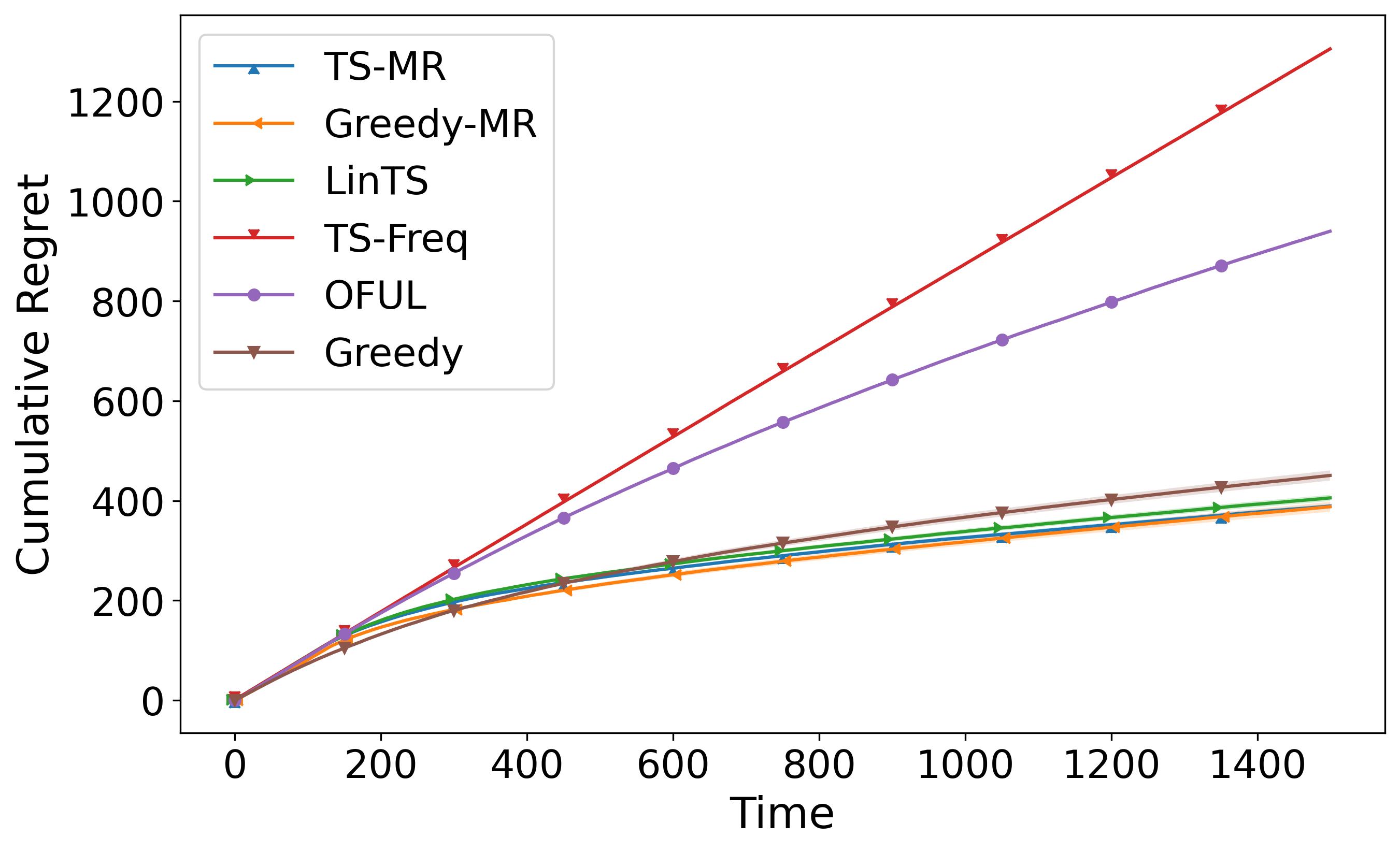} 
	\caption{JapaneseVowels dataset}
		\label{fig:JapaneseVowels}
	\end{subfigure}
 	\begin{subfigure}[b]{0.32\textwidth}
	\includegraphics[width=\textwidth]{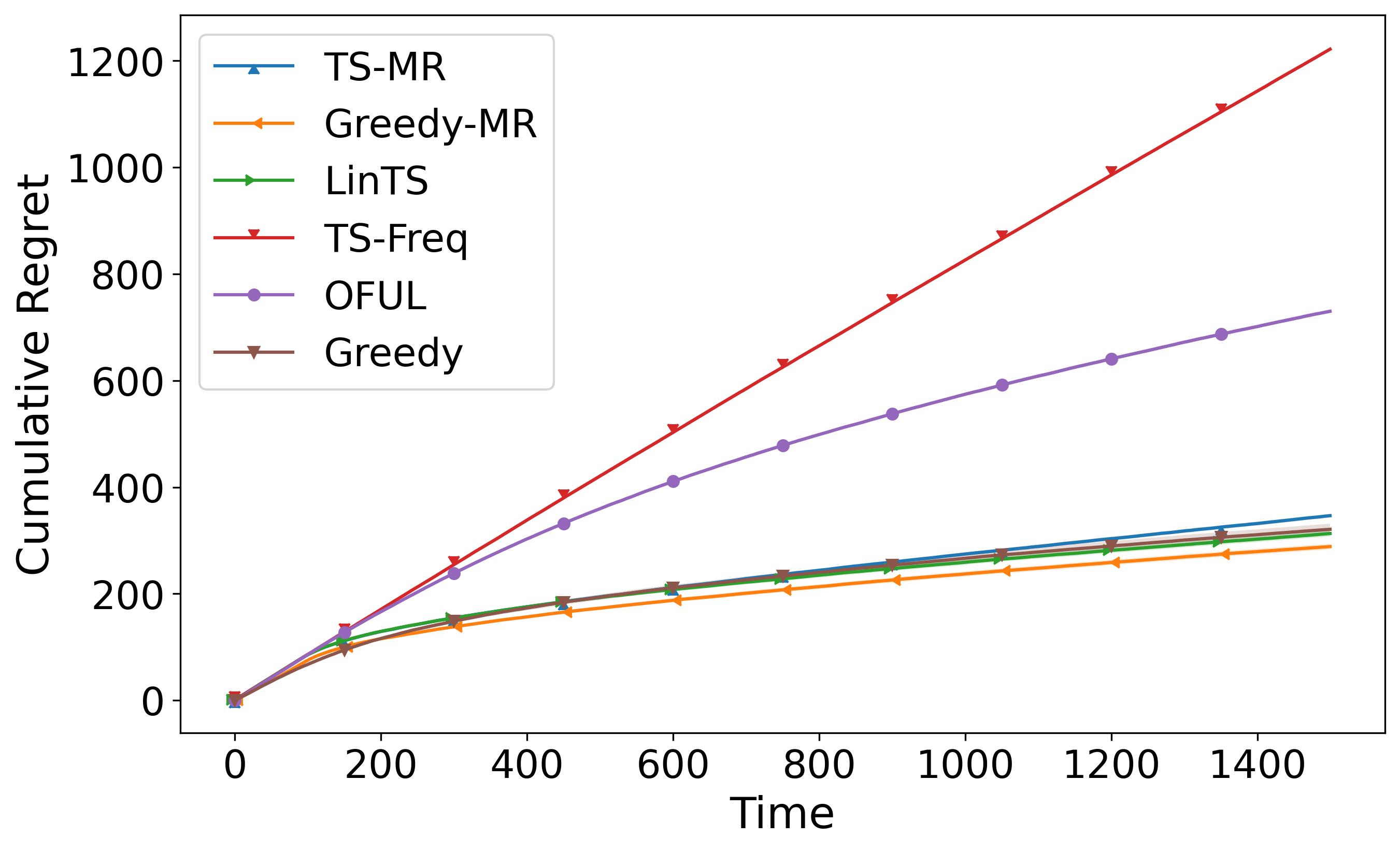} 
	\caption{Segment dataset}
		\label{fig:segment}
	\end{subfigure}\\
    \begin{subfigure}[b]{0.32\textwidth}
		\includegraphics[width=\textwidth]{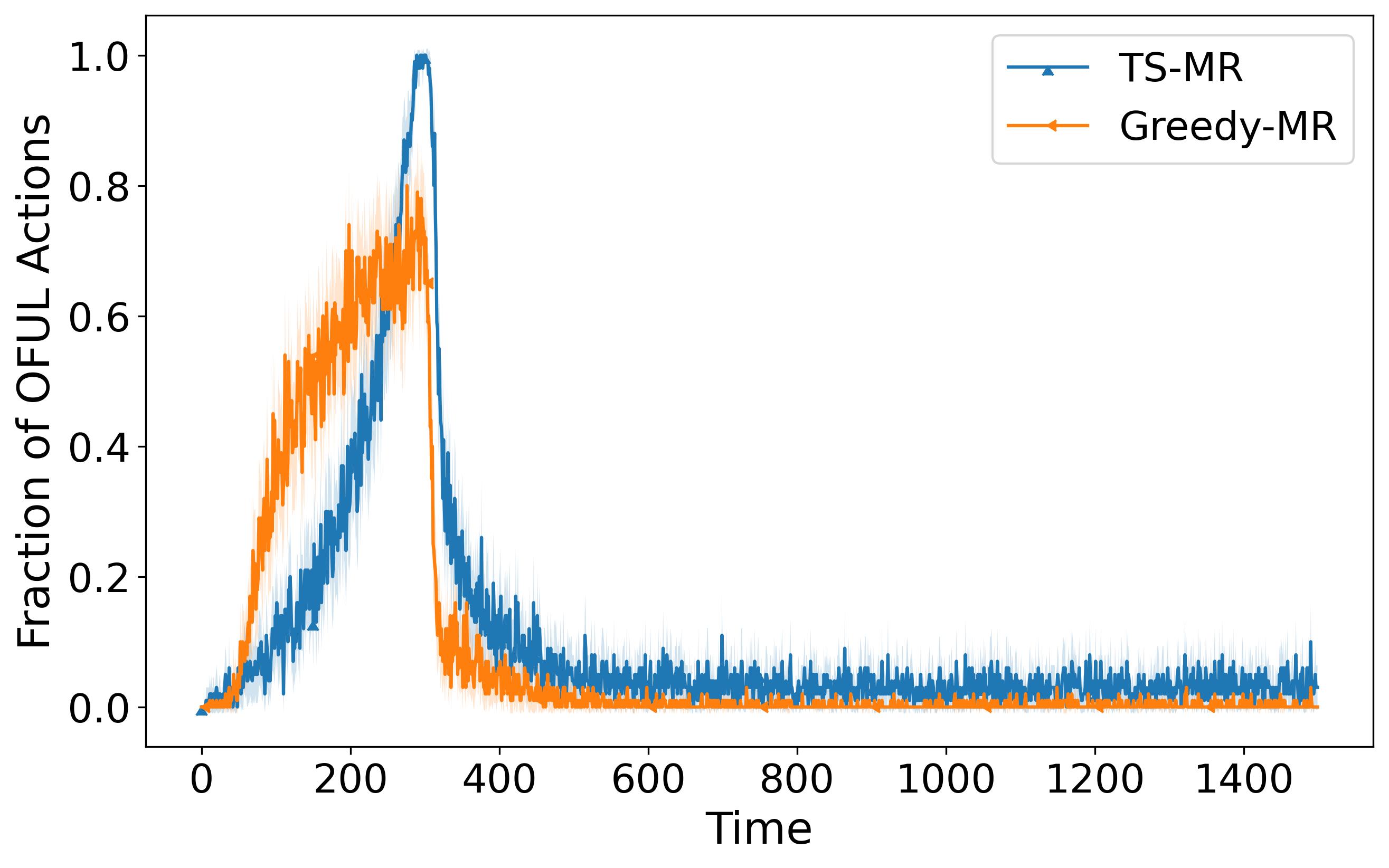} 
	\caption{Cardiotocography dataset}
		\label{fig:Cardiotocography_frac}
\end{subfigure}
	\begin{subfigure}[b]{0.32\textwidth}
	\includegraphics[width=\textwidth]{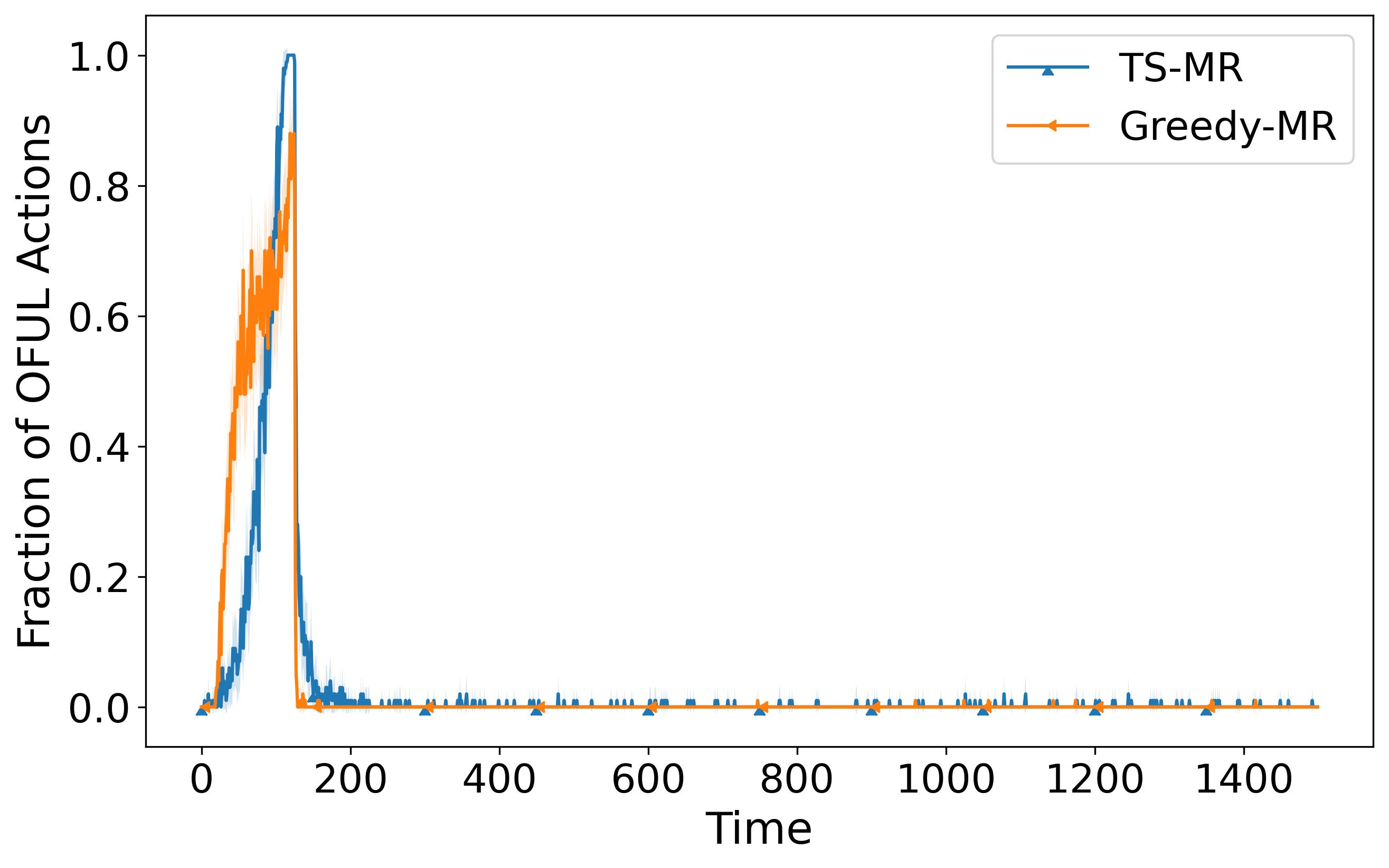} 
	\caption{JapaneseVowels dataset}
		\label{fig:JapaneseVowels_frac}
	\end{subfigure}
 	\begin{subfigure}[b]{0.32\textwidth}
	\includegraphics[width=\textwidth]{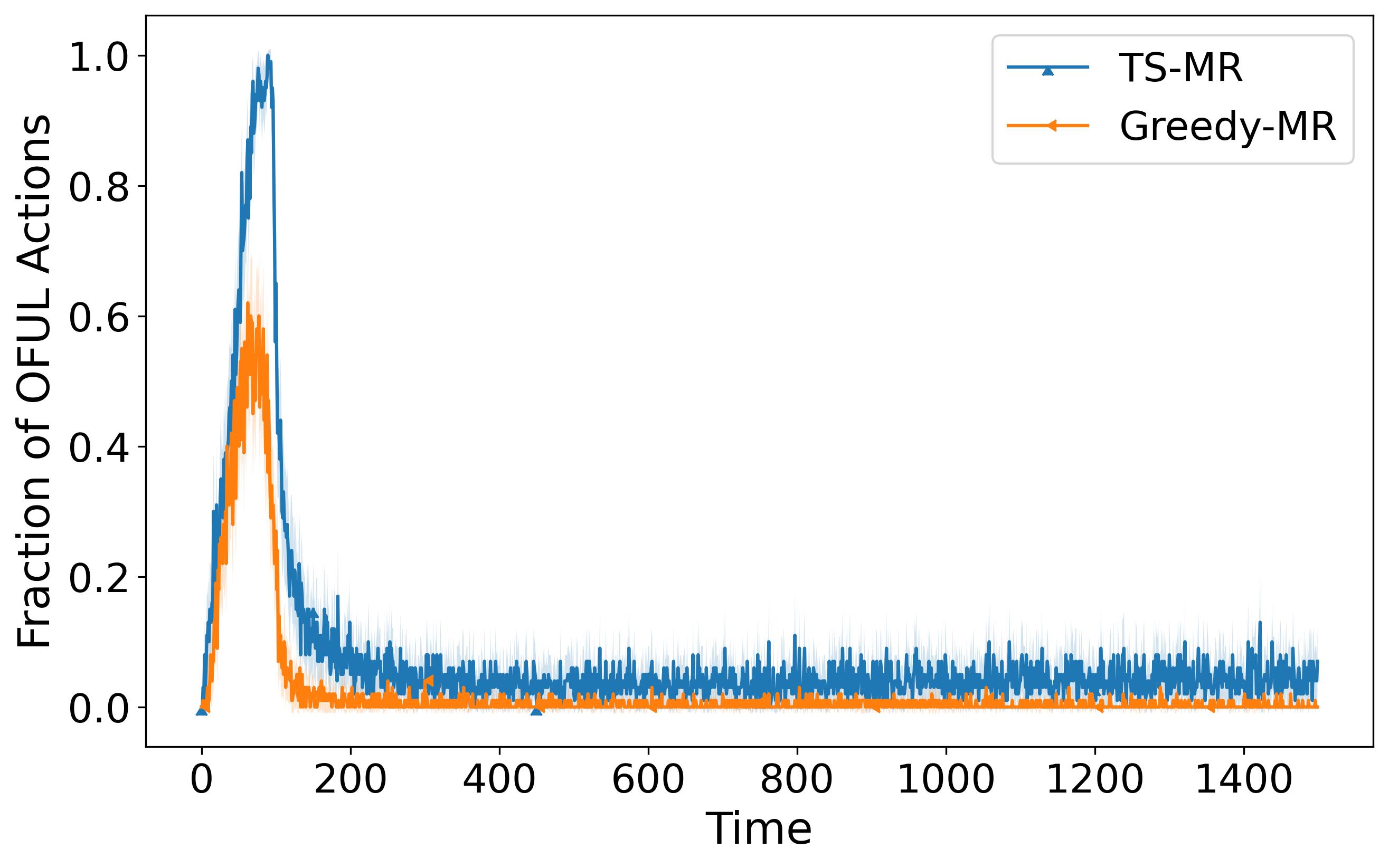} 
	\caption{Segment dataset}
		\label{fig:segment_frac}
	\end{subfigure}
	\caption{Simulation results on real-world datasets. (a) - (c): Cumulative regret of all algorithms. Shaded regions show the $\pm 2$ SE of the mean regret. (d) - (f): Fraction of OFUL actions of TS-MR and Greedy-MR.}
	\label{fig:emp}
\end{figure}

We plot the cumulative regret (averaged over 100 runs) for all algorithms.   Figure~\ref{fig:emp} shows that for all real-world datasets: OFUL and TS-Freq perform poorly due to their conservative exploration; LinTS and Greedy are achieving empirical success even though they don't have theoretical guarantees; TS-MR and Greedy-MR retain the desirable empirical performance of LinTS and Greedy, while enjoying the minimax optimal frequentist regret bound.

\begin{remark}
Simulation results in Figures~\ref{fig:synthetic} and \ref{fig:emp} show OFUL actions are primarily used in the early stages. This indicates: (1) Greedy-MR and TS-MR implement OFUL actions only when necessary, maintaining a low OFUL fraction throughout most of the time horizon, substantially reducing computational cost; and (2) limited course-corrected exploration at the beginning efficiently remedies TS and Greedy in problematic instances.
\end{remark}

\section{Conclusion}
\label{sec:conclusion}

In this work, we propose a data-driven framework to analyze the frequentist regret of \POFUL{}, a family of algorithms that includes OFUL, LinTS, TS-Freq, and Greedy as special cases. Our approach allows for the computation of a data-driven frequentist regret bound for \POFUL{} during implementation, which subsequently informs the course-correction of the algorithm. Our technique conducts a novel real-time geometric analysis of the $d$-dimensional confidence ellipsoid to fully leverage the historical information and might be of independent interest. As applications, we propose TS-MR and Greedy-MR algorithms that enjoy provable minimax optimal frequentist regret and demonstrate their ability to adaptively switch to OFUL when necessary in hard problems where LinTS and Greedy fail. We hope this work provides a steady step towards bridging the gap between theoretical guarantees and empirical performance of bandit algorithms such as LinTS and Greedy.

\begin{APPENDICES}
\section{Sketch of the Proof}
\label{sec:proof_sketch_alpha}
In this section, we present the proof of Theorem~\ref{thm:alpha}.  In the remaining part of this section, we use a general confidence ellipsoid  $\cE_t  \coloneqq \left \{\theta \in \RR^d: \|\theta - \hparam_t\|_{V_t}\leq \beta_t\right \}$ to represent $\cE^{\RLS}_{t,\delta^\prime, \lambdareg}$ and $\cE^{\PVT}_{t,\delta^\prime, \lambdareg}$, since the proof works for both of them.

First note that, when $\|\hparam_t\|_{V_t}  < \beta_t$, the bound in Theorem~\ref{thm:alpha} becomes 
\$\alpha_t\leq \lambda_d^{-\frac{1}{2}}(V_t) /\lambda_1^{-\frac{1}{2}}(V_t) = \sqrt{\lambda_1(V_t)/\lambda_d(V_t)}.\$
This bound holds trivially using the fact that $\lambda_1(V_t) \leq \|x\|^2_{V_t} \leq \lambda_d(V_t) $. This is the case when the data is insufficient and the confidence interval is too large to get a non-trivial upper bound for $\alpha_t$.

In the following, without loss of generality we assume $\|\hparam_t\|_{V_t}^2  \geq \beta_t^2$. The proof decomposes into three steps. In the first two steps, as is illustrated in $\RR^2$ in Figure~\ref{fig:ellipsoid}, we cut out a special hypersurface  $\cH_t$ within  $\cE_t$ and show that for all $\theta \in \cH_t$, the corresponding optimal action $x^\star(\theta)$ has $V_t$-norm bounded from above and below. Note that the set of optimal actions for  $\theta \in \cH_t$ coincides with that for  $\theta \in \cE_t$, we get upper and lower bounds for  $V_t$-norm of all potential actions in the ellipsoid. Next, we show that upper and lower bounds for the $V_t$-norm can be converted into upper and lower bounds for the $V_t^{-1}$-norm by solving a linear programming problem. Hence, we get an upper bound for $\alpha_t$ by calculating the ratio of the upper bound to the lower bound. We sketch the proof below and postpone the detailed proof for all lemmas in this section to Appendix~\ref{sec:proof_alpha}.

\begin{figure}[htp!]
	\centering
		\includegraphics[width=.5\textwidth]{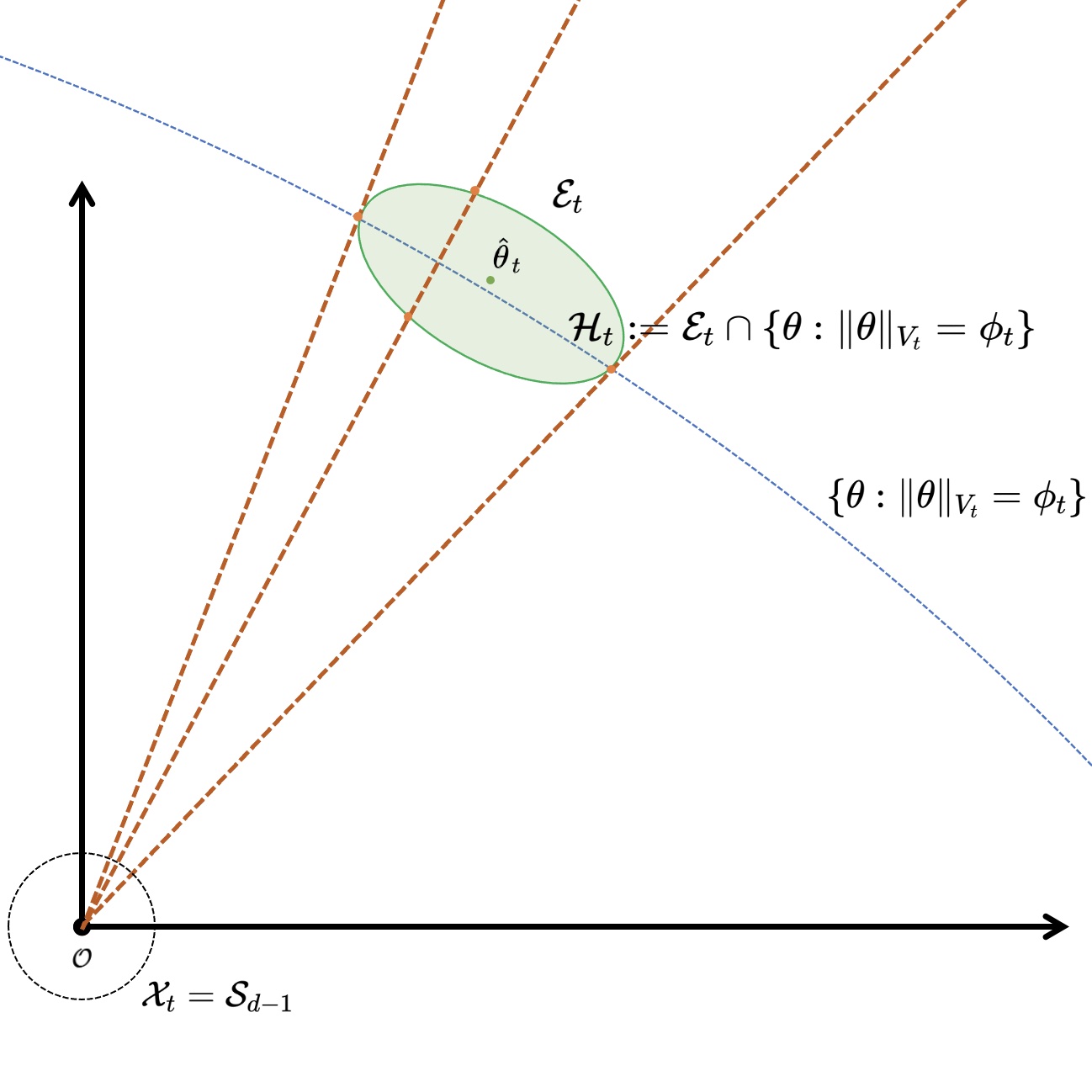}
	\caption{Illustration of Step 1 and 2 in $\RR^2$. Orange dashed rays: rays starting from the origin might have different numbers of intersections with $\cE_t$, indicating whether the corresponding action lies in the projection of $\cE_t$ onto $\cS_{d-1}$. Blue dashed curve:  the ellipsoid with fixed $V_t$-norm  $\left\{\theta:\|\theta\|_{V_t}=\phi_t\right\}$. The intersection of this ellipsoid and $\cE_t$ has the same projection as $\cE_t$ onto $\cS_{d-1}$. }
	\label{fig:ellipsoid}
\end{figure}

In the following, we let
$
\phi_t	 \coloneqq \sqrt{ \|\hparam_t\|_{V_t}^2 - \beta_t^2 }
$, which is well-defined since we assume $\|\hparam_t\|_{V_t}  \geq \beta_t$. In geometry,  one can show that for a ray starting from the origin and intersecting $\cE_t$ only at one point, $\phi_t$ is the ${V_t}$-norm of the intersection point. 

\paragraph{Step 1. Upper Bounding the $V_t$-norm of actions.}
Our first lemma investigates such intersection and provides an upper bound for the  $V_t$-norm for the optimal actions corresponding to any $\theta\in \cE_t$. The proof is based on investigating the condition for a ray paralleling an action $x \in \cS_{d-1}$ to intersect with $\cE_t$, which means $x$ is in the projection of $\cE_t$ onto $\cS_{d-1}$ and might become the optimal action $x^\star(\theta)$.

\begin{lemma}\label{lem:Vt_norm_upper}
For any $\theta\in \cE_t$, we have
	$
	\|x^\star(\theta )\|_{V_t} \leq  \|\hparam_t\|_{V_t^2}/\phi_t$.
\end{lemma}

\paragraph{Step 2. Lower Bounding the $V_t$-norm of actions.}

In order to lower bound the $V_t$-norm, we define the hypersurface  
$
\cH_t \coloneqq \mathcal{E}_t\cap\left\{\theta:\|\theta\|_{V_t}=\phi_t\right\}
$, 
i.e., the intersection of the interior of the confidence ellipsoid  $\cE_t$ and the ellipsoid $\left\{\theta:\|\theta\|_{V_t}=\phi_t\right\}$.
$\cH_t$  consists of $\theta \in \cE_t$ whose $V_t$-norm is $\phi_t$. One can check $\cH_t$ is non-empty since $ \phi_t  \hparam_t  /\|\hparam_t\|_{V_t}  \in \cE_t$, and  the projection of $\cH_t$  onto $\cS_{d-1}$ is the same as that of $\cE_t$ by convexity. Hence,  it suffices to only consider  $\theta \in \cH_t$ as the corresponding set of optimal actions coincides. A lower bound for the $V_t$-norm is given by the following lemma.

\begin{lemma}
\label{lem:Vt_norm_lower}
	For any $\theta\in \cE_t$, we have
	\$
	  \|x^\star(\theta )\|_{V_t} \geq \frac{\phi_t}{\|\hat{\theta}_t\| +  \beta_t/\lambda_d(V_t) }.\$
\end{lemma}
The proof is directly using the fact that for any $ \theta \in \cH_t$, we have $\|\theta\| \leq \|\hparam_t\| +  \beta_t/\lambda_d(V_t)  $ and $\|\theta\|_{V_t}  =  \phi_t$. Also recall  $x^\star(\theta ) = \theta /\|\theta\|$ and hence $\|x^\star(\theta )\|_{V_t} =  \|\theta\|_{V_t} / \|\theta\|$.

\paragraph{Step 3. Bounding the $V_t^{-1}$-norm of actions}

The following lemma determines the range of action $x$'s $V_t^{-1}$-norm based on its $V_t$-norm range. It turns out that the two ranges can be related using the spectral information of the sample covariance matrix $V_t$, which is related to the shape of the confidence ellipsoid.

\begin{lemma}
\label{lem:Vinv_norm}
Let $\{\lambda_1,\lambda_2,\dots, \lambda_{N_V}\}$ be the set of distinct eigenvalues of $V$ such that $\lambda_1>\lambda_2>\dots> \lambda_{N_V}>0$. Let $x\in \cB_d$ satisfies
	$
	0< m \leq \|x\|_{V}^2 \leq M
	$. We have
	\#
	\frac{1}{ \lambda_{k}} + \frac{1}{ \lambda_{k+1}}   - \frac{M}{\lambda_{k}\lambda_{k+1} } \leq \|x\|_{V^{-1}}^2  \leq  \frac{1}{ \lambda_{1}} + \frac{1}{ \lambda_{d}}   - \frac{m}{\lambda_{1}\lambda_{d} },
	\#
	where $k$ is such that $\lambda_{k} = \max_{i \in[N_v]} \{\lambda_i  \geq M\}$.
\end{lemma}

The proof involves expressing the $V$- and $V^{-1}$-norms as weighted sums of $V$'s eigenvalues, then solving a linear programming (LP) problem constrained by the norm ranges.

By inserting the upper and lower bounds of the $V_t$-norm from Lemmas~\ref{lem:Vt_norm_upper} and \ref{lem:Vt_norm_lower} into Lemma~\ref{lem:Vinv_norm}, we finalize the proof of Theorem~\ref{thm:alpha}.

\section{Proof of Lemmas for Theorem~\ref{thm:alpha}}
\label{sec:proof_alpha}
\subsection{Proof of Lemma~\ref{lem:Vt_norm_upper}}

\proof{}
	Let $\theta = t x $ where $x$ is any unit vector in $\RR^d$ and $t\in \RR^{+}  $ is a scalar. Consider the equation that characterizes the intersection  $\{t x : t\in\RR^{+} \}\cap \cE_t $, namely
	$
	(t x - \hparam_t)^\top V_t (t x - \hparam_t) \leq  \beta_t^2
	$.
	Equivalently, we have
	$
	t^2\|x\|_{V_t}^2 - 2 t  x^\top V_t \hparam_t + \phi_t^2 \leq 0
	$. This quadratic inequality of $t$ has at least one solution if the discriminant is non-negative, i.e.
	$
	4(x^\top V_t \hparam)^2 \geq 4 \|x\|_{V_t}^2 \phi_t^2 
	$.
	Then by direct computation,
	\$
	\|x\|_{V_t} \leq \frac{\sqrt{(x^\top V_t \hparam_t)^2}}{\phi_t} \leq \frac{\sqrt{x^\top x}\sqrt{(\hparam_t)^\top V_t^\top V_t \hparam_t }}{\phi_t} =  \frac{ \|\hparam_t\|_{V_t^2}} {\phi_t	}.
	\$

	Note that $\theta \in \cE_t$ if and only if $\theta$ is on a ray starting from the origin and intersects $\cE_t$ at one or more points. Namely, 
	$\theta = t x $ for some $x$ that satisfies $4(x^\top V_t \hparam)^2 \geq 4 \|x\|_{V_t}^2 \phi_t^2$, we conclude the proof.

\endproof
\subsection{Proof of Lemma~\ref{lem:Vt_norm_lower}}
\proof{}
Note that for any $\theta \in \cH_t \subset\cE_t$, it holds that  $\|\theta\| \leq \|\hparam_t\| +  \beta_t/\lambda_d(V_t)  $. Also, by the construction of $\cH_t$, we have $\|\theta\|_{V_t}  =  \phi_t$. Then by direct computation, we have
	\$
	\|x^\star(\theta )\|_{V_t} = \frac{\|\theta\|_{V_t}}{\|\theta\| } \geq \frac{\phi_t}{\|\hat{\theta}_t\| +  \beta_t/\lambda_d(V_t) } .
	\$

To prove the same result for any $\theta \in \cE_t$, we only need to show there exists $\theta^\prime \in \cH_t$ such that $x^\star(\theta) = x^\star(\theta^\prime)$. To see this,  let $x = \theta/\|\theta\|$ and consider the  intersection  $\{t x : t\in\RR^{+}  \}\cap \cE_t$, which is non-empty by our choice of $\theta$. Similar to the proof of Lemma \ref{lem:Vt_norm_upper}, the discriminant is non-negative, i.e.
	$
	4(x^\top V_t \hparam)^2 \geq 4 \|x\|_{V_t}^2 \phi_t^2 
	$. 

Now consider the intersection  $\{t x : t\in\RR \}\cap \partial\cE_t$, where we let $\partial \cE_t \coloneqq \left \{\theta \in \RR^d: \|\theta - \hparam_t\|_{V_t}= \beta_t(\delta)\right \}$ be the border of the ellisoid. The intersection points are characterized by the solution to 
\#\label{eq:border_intersection}
t^2\|x\|_{V_t}^2 - 2 t  x^\top V_t \hparam_t + \phi_t^2 = 0.
\#

If $
	4(x^\top V_t \hparam)^2 = 4 \|x\|_{V_t}^2 \phi_t^2 
	$ and there is only one intersection point, namely $\theta$ itself, we have
	\$
	0 = t^2\|x\|_{V_t}^2 - 2 t  x^\top V_t \hparam_t + \phi_t^2 = t^2\|x\|_{V_t}^2 - 2 t  \|x\|_{V_t} \phi_t  + \phi_t^2 = (\|tx\|_{V_t} - \phi_t)^2.
	\$
Therefore, we have $\|\theta\|_{V_t} = \|tx\|_{V_t} = \phi_t$, i.e.  $\theta \in \cH_t$.

If $4(x^\top V_t \hparam)^2 > 4 \|x\|_{V_t}^2 \phi_t^2$, it follows that $x^\top V_t \hparam > \|x\|_{V_t}\phi_t$. This inference is valid given that $x^\top V_t \hparam > 0$, which can be verified using Equation \eqref{eq:border_intersection} and noting that $t>0$.
Consider the solutions to \eqref{eq:border_intersection}
\$
t_1 = \frac{x^\top V_t \hparam_t - \sqrt{(x^\top V_t \hparam)^2 - \|x\|_{V_t}^2 \phi_t^2 }  }{\|x\|_{V_t}^2}, \quad t_2 = \frac{x^\top V_t \hparam_t + \sqrt{(x^\top V_t \hparam)^2 - \|x\|_{V_t}^2 \phi_t^2 }  }{\|x\|_{V_t}^2}.
\$
We only need to show $ \|t_1 x\|_{V_t}< \phi_t  < \|t_2 x\|_{V_t}$, then by the continuity of $\|\cdot\|_{V_t}$ and the convexity of $\cE_t$, there exists $t^\prime \in (t_1,t_2)$ such that $\|t^\prime x\|_{V_t} = \phi_t $. Then $\theta^\prime \coloneqq t^\prime x \in \cH_t$ is the desired point. By direct computation, 
\$
\|t_1 x\|_{V_t} = \frac{x^\top V_t \hparam_t - \sqrt{(x^\top V_t \hparam)^2 - \|x\|_{V_t}^2 \phi_t^2 }  }{\|x\|_{V_t}}.
\$
We only need to prove 
\$
x^\top V_t \hparam_t -\|x\|_{V_t}  \phi_t  \leq \sqrt{(x^\top V_t \hparam)^2 - \|x\|_{V_t}^2 \phi_t^2 } 
\$
Note that 
\$
(x^\top V_t \hparam_t -  \|x\|_{V_t}\phi_t ) ^2  &=  (x^\top V_t \hparam)^2  -2x^\top V_t \hparam \phi_t \|x\|_{V_t} + \|x\|_{V_t}^2 \phi_t^2 \\
&\leq (x^\top V_t \hparam)^2  - 2 \|x\|_{V_t}^2 \phi_t^2 + \|x\|_{V_t}^2 \phi_t^2\\
& = (x^\top V_t \hparam)^2  -  \|x\|_{V_t}^2 \phi_t^2
\$
where we have used the fact that $x^\top V_t \hparam > \|x\|_{V_t}\phi_t$ in the ineuqlity.  Taking square root for both sides yields the desired result, and hence $ \|t_1 x\|_{V_t}  < \phi_t$. Similarly, one can show $\phi_t  < \|t_2 x\|_{V_t}$. This concludes the proof.

\endproof

\subsection{Proof of Lemma~\ref{lem:Vinv_norm}}

\proof{}
	Let $\{\lambda_1,\lambda_2,\dots, \lambda_{N_V}\}$ be the set of distinct eigenvalues of $V$.  By the spectral theorem, $V$ can be decomposed as $V = U \Lambda U^\top  $, where the columns of $U$ consist of orthonormal eigenvectors of $V$, denoted by $\{u_{11},\dots,u_{1 n_1},u_{21}, \dots,u_{2n_2},\dots,u_{N_V 1},\dots,u_{N_V n_{N_V}}\}$, where $n_1,\dots, n_{N_V}$ are the  algebraic multiplicity of the eigenvalues respectively. Since $V$ is a symmetric matrix, the eigenvectors form a basis of $\RR^d$ and we have  $\sum_{i = 1}^{N_V}n_i = d$. 	  We can write $x$ as a linear combination 
	$
	x = \sum_{i=1}^{N_V} \sum_{j=1}^{n_i} w_{ij}  u_{ij}
	$,
	where
	$
	\sum_{i=1}^{N_V} \sum_{j=1}^{n_i} w_{ij}^2 = 1
	$. Define $a_i \coloneqq \sum_{j=1}^{n_i} w_{ij}^2 $, by direct computation, we have $ \sum_{i=1}^{N_V} a_i = 1 $  and $\|x\|_V^2 = \sum_{i=1}^{N_V} \lambda_ia_i $ and  $\|x\|_{V^{-1}}^2 = \sum_{i=1}^{N_V} \lambda_i^{-1} a_i $ .

	Now we study the range of $\|x\|_{V^{-1}}^2$ when $\|x\|_{V}^2$ is bounded as $m \leq \|x\|_{V}^2 \leq M$. First, let's focus on  maximizing $\|x\|_{V^{-1}}^2$, it suffices to solve the LP problem
	\$
	\text{maximize}& \sum_{i=1}^{N_V} a_i\lambda_i^{-1} \quad
	\text{s.t.} \quad \forall i, a_i\geq 0, \ 
	\sum_{i=1}^{N_V}  a_i = 1, \ 
	\sum_{i=1}^{N_V} a_i \lambda_i \geq m,\ 
	\sum_{i=1}^{N_V}  a_i \lambda_i \leq M.
	\$
	The Lagrangian is given by
	\$
	L = \sum_{i=1}^{N_V}  a_i\lambda_i^{-1} + \mu ( 1- \sum_{i=1}^{N_V}  a_i) + \eta (\sum_{i=1}^{N_V}  a_i \lambda_i - m)+ \gamma ( M - \sum_{i=1}^{N_V}  a_i \lambda_i )+\sum_{i=1}^{N_V}  \kappa_ia_i .
	\$
	The KKT conditions are given by
	\#\begin{cases}\label{eq:KKT}
		 & \nabla_{a_i} L  = \lambda_i^{-1} - \mu + \eta \lambda_i  -  \gamma \lambda_i+ \kappa_i =0, \forall i,                         \\
		 & a_i\geq 0, \forall i, \sum_{i=1}^{N_V}  a_i = 1, \sum_{i=1}^{N_V} a_i \lambda_i \geq m, \sum_{i=1}^{N_V}  a_i \lambda_i \leq M,  \\
		 & \eta \geq  0,\gamma \geq 0,\kappa_i \geq 0, \forall i,                                                             \\
		 & \eta(\sum_{i=1}^{N_V} a_i \lambda_i - m) = 0, \gamma(M - \sum_{i=1}^{N_V} a_i \lambda_i) = 0,\kappa_i a_i = 0,  \forall i.
	\end{cases}
	\#

	To satisfy the first condition above, $\kappa_i$'s can only be zero for at most two indices.  Hence, $a_i$ can only be non-zero for at most two distinct eigenvalues, denoted by $\lambda_i$ and $\lambda_j$, where  $i<j$ and  $\lambda_i >\lambda_j$. Namely, the solution to  \eqref{eq:KKT}  lies in the subspace spanned by the eigenvectors corresponding to $\lambda_i$ and $\lambda_j$.

	Let $y = \|x\|_V^2 $, we have $a_i\lambda_i + a_j \lambda_j =y$ and $a_i\lambda_i^{-1} + a_j \lambda_j ^{-1} = \|x\|_{V^{-1}}^2$. Note that $\sum_{i=1}^{N_V} a_i = a_i + a_j  = 1$, by direct computation, the closed form of $\|x\|_{V^{-1}}^2 $ is given by
	$
	\|x\|_{V^{-1}}^2 = \frac{1}{ \lambda_{i}} + \frac{1}{ \lambda_{j}}   - \frac{y}{\lambda_{i}\lambda_{j} }\eqqcolon f(y,\lambda_{i},\lambda_{j} )
	$. 
	
	Clearly, we have
	\$
	\begin{cases}
		\frac{\partial f}{\partial y}           & = -\frac{1}{\lambda_{i}\lambda_{j} } < 0,              \\
		\frac{\partial f}{\partial \lambda_{i}} & = (\frac{y}{\lambda_j} - 1)\frac{1}{\lambda_i^2 } > 0, \\
		\frac{\partial f}{\partial \lambda_{j}} & = (\frac{y}{\lambda_i} - 1)\frac{1}{\lambda_j^2 } < 0.
	\end{cases}
	\$

	Then the maximum of $\|x\|_{V^{-1}}^2 $ is obtained when $\lambda_i = \lambda_1$, $\lambda_j = \lambda_{N_V}$, $y = m$. Therefore, the solution to the LP problem is any unit vector $x^{\star}_{\max}$ that lies in the subspace spanned by the eigenvectors corresponding to $\lambda_1$ and $\lambda_{N_V}$. Moreover, we have
	\$
	\|x^\star_{\max}\|_{V^{-1}}^2 = \frac{1}{ \lambda_{1}} + \frac{1}{ \lambda_{N_V}}   - \frac{m}{\lambda_{1}\lambda_{N_V} }.
	\$

	Similarly, by investigating the KKT conditions for the LP problem that minimize $\sum_{i=1}^{N_V} a_i\lambda_i^{-1}$,   the minimum of $\|x\|_{V^{-1}}^2 $ is obtained when  $\lambda_i = \lambda_{k} $, $\lambda_j = \lambda_{k+1}$, $y = M$, where $k$ is such that $\lambda_{k} = \max_{i \in[N_v]} \{\lambda_i  \geq M\}$, and hence $\lambda_{k+1} = \min_{i \in[N_v]} \{\lambda_i  < M\}$. The solution vector is  any unit vector $x^{\star}_{\min}$ that lies in the subspace spanned by the eigenvectors corresponding to $\lambda_{k}$ and $\lambda_{k+1}$, and we have
	\$
	\|x^\star_{\min}\|_{V^{-1}}^2 = \frac{1}{ \lambda_{k}} + \frac{1}{ \lambda_{k+1} }   - \frac{M}{\lambda_{k}\lambda_{k+1}  }.
	\$
	This concludes the proof.

\endproof

\section{Other Proofs}
\label{sec:other_proofs}
\subsection{Proof of Proposition~\ref{prop:concentration} }
\label{proof:concentration}
\proof{}
	\label{pf:prop_rls}
	By Proposition~\ref{prop:beta}, we have
	\$
	\prob{\hat{\cA}_T}
	& = \prob{\cap_{t = 1}^T \left\{ \|\hparam_t - \param_t\|_{V_t} \geq \beta_{t,\delta^\prime, \lambdareg}^{RLS} \right\}
	}  \\
	&\geq 1 - \sum_{t = 1}^T \prob{ \|\hparam_t - \param_t\|_{V_t}\geq\beta_{t,\delta^\prime, \lambdareg}^{RLS} }\\
	&\geq 1 - \frac{\delta}{2}.
	\$
	
	Similarly, by Definition~\ref{def:tilde_beta}, we have 
		\$
	\prob{\tilde{\cA}_T}
	& = \prob{\cap_{t = 1}^T \left\{ \tilde{\theta}_t \notin\cE_{t,\delta^\prime, \lambdareg}^{\PVT} |\mathcal{F}_t\right\}
	}  \\
	&\geq 1 - \sum_{t = 1}^T \prob{ \tilde{\theta}_t \notin \cE_{t,\delta^\prime, \lambdareg}^{\PVT} |\mathcal{F}_t }\\
	&\geq 1 - \frac{\delta}{2}.
	\$
Combining the two inequalities above, we have $\prob{\cA_T} \geq 1 - \delta$. 
\endproof

\subsection{Proof of Proposition~\ref{prop:instantaneous} }
\label{proof:instantaneous}
\proof{}Since ${\theta}^\star \in \cE_{t,\delta^\prime, \lambdareg}^{RLS} $ and $\tilde{\theta}_t \in \cE_{t,\delta^\prime, \lambdareg}^{\PVT}$, it holds that 
\#\label{eq:two_radius}
 \|{\theta}^\star - \hparam_t\|_{V_t}\leq \beta_{t,\delta^\prime, \lambdareg}^{RLS},  \quad \|\tilde{\theta}_t - \hparam_t\|_{V_t}\leq \beta^{\PVT}_{t,\delta^\prime, \lambdareg} = \iota_t \beta_{t,\delta^\prime, \lambdareg}^{RLS}.
\#


We have 
\$
 \dotp{x_t^\star}{ \theta^{\star}} - \dotp{\tilde{x}_t}{ \theta^{\star}}
& = \left(\dotp{x^{\star}_t}{ \theta^{\star}} - \dotp{x^{\star}_t}{ \tilde{\theta}_t}\right) + \left(\dotp{x^{\star}_t}{ \tilde{\theta}_t} - \dotp{\tilde{x}_t}{  \tilde{\theta}_t}\right)\\
& \quad + \left(\dotp{\tilde{x}_t}{\tilde{\theta}_t}-\dotp{\tilde{x}_t}{ \hparam_t}\right) +  \left(\dotp{\tilde{x}_t}{\hparam_t}-\dotp{\tilde{x}_t}{ \theta^{\star}}\right).
\$

To bound the second term on the right hand side, recall by Equation \eqref{eq:POFUL} the \POFUL{} action $\tilde{x}_t$ satisfies 
\$
\dotp{x^{\star}_t}{ \tilde{\theta}_t} +  \tau_t  \|x^{\star}_t\|_{V^{-1}_t}\beta_{t,\delta^\prime, \lambdareg}^{RLS} \leq  \dotp{\tilde{x}_t}{  \tilde{\theta}_t} + \tau_t  \|\tilde{x}_t\|_{V^{-1}_t}\beta_{t,\delta^\prime, \lambdareg}^{RLS}.
\$
Rearranging the ineuqlity, we obtain $\dotp{x^{\star}_t}{ \tilde{\theta}_t} - \dotp{\tilde{x}_t}{  \tilde{\theta}_t} \leq \tau_t\|\tilde{x}_t\|_{V^{-1}_t} \beta_{t,\delta^\prime, \lambdareg}^{RLS}-\tau_t \|x^{\star}_t\|_{V^{-1}_t} \beta_{t,\delta^\prime, \lambdareg}^{RLS}$.

The other three terms are bounded similarly using the Cauchy-Schwarz inequality,  the triangle inequality of the $V_t^{-1}$-norm,  and the concentration condition \eqref{eq:two_radius}. As a result, we have
\$
\dotp{x^{\star}_t}{ \theta^{\star}} - \dotp{x^{\star}_t}{ \tilde{\theta}_t} &\leq (1+\iota_t)\|x^{\star}_t\|_{V^{-1}_t}\beta_{t,\delta^\prime, \lambdareg}^{RLS} , \\ \dotp{\tilde{x}_t}{\tilde{\theta}_t}-\dotp{\tilde{x}_t}{ \hparam_t}  &\leq  \iota_t \|\tilde{x}_t\|_{V^{-1}_t}\beta^{RLS}_{t,\delta^\prime, \lambdareg},\\
\dotp{\tilde{x}_t}{\hparam_t}-\dotp{\tilde{x}_t}{ \theta^{\star}} &\leq  \|\tilde{x}_t\|_{V^{-1}_t}\beta^{RLS}_{t,\delta^\prime, \lambdareg}.
\$
Combining all terms above, we have 
\$
\dotp{x_t^\star}{ \theta^{\star}} - \dotp{\tilde{x}_t}{ \theta^{\star}}
 \leq  (1+\iota_t - \tau_t)\|x^{\star}_t\|_{V^{-1}_t}\beta_{t,\delta^\prime, \lambdareg}^{RLS} + (1+\iota_t + \tau_t) \|\tilde{x}_t\|_{V^{-1}_t}\beta^{RLS}_{t,\delta^\prime, \lambdareg}.
\$
\endproof

\subsection{Proof of Theorem~\ref{thm:oracle_regret}}
\label{sec:proof_regret}
\proof{}
We formally prove Theorem~\ref{thm:oracle_regret} for completeness. The proof techniques are developed in previous papers \citep{abbasi2011improved,agrawal2013thompson,abeille2017linear}.

Throughout the proof, we condition on the event $\cA_T$, which holds with probability $1-\delta$ by Proposition~\ref{prop:concentration}. Applying Proposition~\ref{prop:instantaneous}, we obtain
\$
\cR(T) 
& \leq \sum_{t = 1}^T \left( \dotp{x_t^\star}{\theta^{\star}}-\dotp{\tilde{x}_t}{\theta^{\star}} \right)\event{\cA_t}\\
& \leq  \sum_{t = 1}^T  (1+\iota_t-\tau_t)\|x^{\star}_t\|_{V^{-1}_t}\beta_{t,\delta^\prime, \lambdareg}^{RLS} +   (1+\iota_t+\tau_t)\|\tilde{x}_t\|_{V^{-1}_t}\beta_{t,\delta^\prime, \lambdareg}^{RLS}.
\$

Recall $\|x^{\star}_t\|_{V^{-1}_t} = \alpha_t \|\tilde{x}_t\|_{V^{-1}_t}$ and $\mu_t  =  \alpha_t(1+\iota_t -\tau_t)+1+\iota_t+\tau_t$, we have 
\$
\cR(T)  \leq \sum_{t = 1}^T   \mu_t \|\tilde{x}_t\|_{V^{-1}_t}\beta_{t,\delta^\prime, \lambdareg}^{RLS}.
\$ 

Applying the Cauchy-Schwarz inequality and Proposition~\ref{prop:potential}, note that $\max_{t\in[T]}  \beta_{t,\delta^\prime, \lambdareg}^{RLS} = \beta_{T,\delta^\prime, \lambdareg}^{RLS}$, we obtain 
\$
\cR(T) &\leq   \sqrt{ \sum_{t=1}^T \|\tilde{x}_t\|_{V_t^{-1}} ^2 } \sqrt{\sum_{t=1}^T\mu_t^2 \left(\beta_{t,\delta^\prime, \lambdareg}^{RLS}\right)^2} \leq \sqrt{2 d \log \left(1+\frac{T}{\lambda}\right)}   \sqrt{\sum_{t=1}^T\mu_t^2 } \beta_{T,\delta^\prime, \lambdareg}^{RLS}.
\$

This concludes the proof.
\endproof

\endproof

\section{Regret Decomposition of \POFUL{}}
\label{sec:regret_decomp}
To discuss the relationship between our method and those based on optimism \citep{abbasi2011improved,agrawal2013thompson,abeille2017linear}, we decompose the regret of \POFUL{} into three terms,  and sketch how they are bounded separately. 

Let  $x_t^\star = \argmax_{x\in \cX_t} \dotp{x}{\theta^\star}$ be the optimal action and $\tilde{x}_t$ be the action chosen by \POFUL{}, the regret decomposes as
\begin{align*}
\cR(T) & =\sum_{t=1}^T\dotp{x_t^\star}{ \theta^{\star}}-\dotp{\tilde{x}_t}{\theta^{\star}}\\
& = \underbrace{\sum_{t=1}^T\dotp{x_t^\star}{ \theta^{\star}}-\dotp{\tilde{x}_t}{\tilde{\theta}_t}}_{ \cR^{\PE}(T)}+ \underbrace{\sum_{t=1}^T\dotp{\tilde{x}_t}{\tilde{\theta}_t}-\dotp{\tilde{x}_t}{ \hparam_t}}_{\cR^{\PVT}(T)} +  \underbrace{\sum_{t=1}^T\dotp{\tilde{x}_t}{\hparam_t}-\dotp{\tilde{x}_t}{ \theta^{\star}}}_{\cR^{\RLS}(T)}\,.
\end{align*}
$\cR^{RLS}(T)$ is the regret due to the estimation error of the RLS estimator. Note that $\tilde{x}_t$'s are the action sequence used to construct the RLS estimate. By Proposition~\ref{prop:potential} their cumulative $V_t^{-1}$-norm is bounded by $2d\log(1+t/\lambdareg)$. Hence, the upper bound of $\cR^{RLS}(T)$ is essentially determined by the $V_t$-distance between $\hparam_t$ and $\param$, which, via the Cauchy-Schwarz inequality, is characterized by the radius of the confidence ellipsoid $\cE^{RLS}_{t,\delta^\prime, \lambdareg}$. $\cR^{\PVT}(T)$ corresponds to the regret due to the exploration of \POFUL{} by choosing the pivot parameter $\tilde{\theta}_t$ rather than using $\hat\theta_t$ as OFUL would. Similar to  $\cR^{RLS}(T)$, the upper bound for this term is related to the $V_t$-distance between $\tparam_t$ and $\hparam_t$, which is controlled by construction and depends on the inflation parameter $\iota_t$. As a result, it can be shown that 
$\cR^{RLS}(T)  = \tilde{\cO}(d\sqrt{T}) $ and $\cR^{\PVT}(T)  = \tilde{\cO}(d(\sum_{t = 1}^T \iota_{t}^2)^{1/2}) $ with probability $1-\delta$.  Notably, $\cR^{RLS}(T) = \tilde{\cO}(d\sqrt{T})  $, matching the known minimax lowerbound. For $\cR^{\PVT}(T)$ to match this lower bound as well, one way is to set $\iota_{t}=
\tilde{\cO}(1)$ for all $t\in[T]$. However, it is known that, see for example \citep{agrawal2013thompson,abeille2017linear,hamidi2020frequentist}, such selection could be problematic for bounding $\cR^{PE}(T)$.

$\cR^{PE}(T)$ corresponds to the \emph{pessimism} term. In the Bayesian analysis of LinTS, the expectation of this term is $0$, with respect to $\theta^\star$, as long as $\tparam_t$ and $\param$ are sampled from the same distribution, which occurs when LinTS has access to the true prior distribution for $\theta^\star$. In the frequentist analysis, however, we need to control the pessimism term incurred by any random sample $\tparam_t$.

For OFUL, this term is bounded properly with high probability by the optimistic selection of OFUL actions\footnote{The term is further split into two parts as shown in the first few lines of proof of Proposition \ref{prop:instantaneous} in Appendix~\ref{proof:instantaneous}, with one term bounded by the optimism and the other term controlled by the $V_t$ distance between $\theta^{\star}$ and $\tilde{\theta}_t$.}. For LinTS, the only known analysis due to \cite{agrawal2013thompson} and \cite{abeille2017linear} gives a bound of order $\tilde{\cO}(d\sqrt{dT})$ using an optimism-based approach, which is worse than the Bayesian regret by a factor of $\sqrt{d}$. The key component of their proof is introducing inflation to the posterior variance by setting $\iota_t =\tilde{\cO}(\sqrt{d})$ to enforce exploration. For example, \cite{abeille2017linear} demonstrate that by using inflation, LinTS samples optimistic $\tparam_t$ with a probability greater than a constant, which means that the inequality $\dotp{x_t^\star}{ \theta^{\star}}\leq \dotp{\tilde{x}_t}{\tilde{\theta}_t}$ holds with a constant probability.

For Greedy, there is no general theoretical guarantee for bounding the pessimism term without imposing additional structural assumptions. Our approach to tackle this challenge distinguishes us from methods based on optimism \citep{agrawal2013thompson,abeille2017linear}. Interestingly, \cite{abeille2017linear} conjectured that non-optimistic samples may provide beneficial exploration and help control the growth of regret.

By relaxing the requirement for optimistic samples, we avoid inflating the posterior, which in turn prevents a $\sqrt{d}$-gap in the regret. Instead, we introduce a measure of the ``quality'' of \POFUL{} actions with respect to the regret and develop a computable upper bound for this quality using geometric information in the data. This approach yields a data-driven regret bound for \POFUL{} that matches the minimax optimal regret in some settings. Furthermore, it allows for the construction of variants of LinTS and Greedy with a provable frequentist regret bound that is minimax optimal up to only a constant factor.

\section{Discrete Action Space}
\label{sec:discrete_actions}


This section introduces the method for establishing an upper bound on $\alpha_t$ in the context of discrete action spaces.  It turns out that the process involves comparing confidence interval lengths across all potentially optimal actions.

To see this, recall for an arbitrary action $x\in\cX_t$,  by Proposition~\ref{prop:beta}, with high probability its expected reward is bounded as 
\$
L_t(x) \coloneqq \dotp{x}{ \hat{\theta}_t}  - \iota_t \|x\|_{V_t^{-1}}   \beta_{t,\delta^\prime, \lambdareg}^{RLS}       
\leq  \dotp{x}{{\theta}^\star}   \leq \dotp{x}{ \hat{\theta}_t}  +  \iota_t \|x\|_{V_t^{-1}}   \beta_{t,\delta^\prime, \lambdareg}^{RLS} \eqqcolon  U_t(x).
\$ 
The confidence interval length is proportional to $\|x\|_{V_t^{-1}} $. Therefore, to characterize the range of $\|x_t^\star\|_{V_t^{-1}}$, we only need to identify the set of potentially optimal actions and calculate the minimal $V_t^{-1}$-norm among all actions in that set.

Note that for an action $x \in \cX_t$  to remain potentially optimal, it cannot be dominated by another action. I.e.,    we require  $U_t(x) \geq \max_{y\in\cX_t}  L_t(y)$. Therefore, define \$\cC_t (\beta)\coloneqq \{x\in\cX_t: \dotp{x}{ \hat{\theta}_t}  +  \|x\|_{V_t^{-1}}  \beta \geq \max_{y\in\cX_t} \  \dotp{y}{ \hat{\theta}_t}  - \|y\|_{V_t^{-1}}    \beta \},\$
the set of potentially optimal actions is given by $C_t(\beta_{t,\delta^\prime, \lambdareg}^{RLS})$. Similarly,  the set of actions that might be chosen by \POFUL{} is given by $\cC_t(\beta_{t,\delta^\prime, \lambdareg}^{\PVT} )$, where we let $\beta^{\PVT}_{t,\delta^\prime, \lambdareg} \coloneqq \iota_t \beta^{RLS}_{t,\delta^\prime, \lambdareg}$. The inflation parameter $\iota_t$ turns out to control the conservativeness when eliminating actions.

By comparing the range of $V_t^{-1}$-norm for both sets, an upper bound $\hat{\alpha}_t$ for the  uncertainty ratio $\alpha_t$ is given by
  \# \label{eq:alpha_hat_2}
  \hat{\alpha}_t\coloneqq \frac{\sup _{x\in \cC_t(\beta_{t,\delta^\prime, \lambdareg}^{RLS} )}\|x\|_{V_t^{-1}}}{ \inf _{x \in \cC_t(\beta_{t,\delta^\prime, \lambdareg}^{\PVT} )}\|x\|_{V_t^{-1}}}.
 \#

\section{Example Cases and empirical validations of Theorem~\ref{thm:alpha}}
\label{sec:Cases}

To better understand what Theorem \ref{thm:alpha} implies, we discuss some special cases in this section. 

\paragraph{Case 1: a pure exploration regime.} \label{eg:pure} When the decision-maker doesn't care about the regret and adopts a pure exploration algorithm that plays actions in all directions sequentially, we expect $\lambda_i(V_t) = \cO(t)$ for all $i = 1,\dots, d$. Then $\hat{\alpha}_t \leq C$ for some constant $C>0$. Specifically, if $V_t = D I_d$ for some constant $D>0$, we have $\|x\|_{V_t^{-1}} =D^{-1} \|x\| = D^{-1}$ for all $x \in \cS_{d-1}$ and hence $ \hat{\alpha}_t = 1$  .

\paragraph{Case 2:  linear structures.} 
\label{eg:principle_eigen}

When the reward takes a linear form in the action, for example, the stochastic linear bandits we study,  it's observed in practice that the estimate $\hat{\theta}_t$ tends to align in the first eigenspace of the uncertainty structure $V_t$ (empirically validated below). To see this, note that in order to maximize the reward in an online manner, bandit algorithms tend to select actions towards the confidence ellipsoid centered at  $\hat{\theta}_t$, hence forcing more exploration along the direction of $\hat{\theta}_t$, especially at the late stage of the algorithm when the ellipsoid is small. The following proposition states that, as long as the center $\hparam_t$ of the confidence ellipsoid $\cE_{t,\delta^\prime, \lambdareg}$  tends to be aligned with the first eigenspace of $V_t$, the corresponding uncertainty ratio tends to 1, indicating a regret bound matching the minimax $\tilde{\cO}(d\sqrt{T})$ rate by Theorem~\ref{thm:oracle_regret}. This provides a plausible explanation for the empirical success of LinTS and Greedy.
\begin{proposition}\label{prop:align}
	 Suppose $\lim_{t\to \infty}\frac{\|\hparam_t\|_{V_{t}}^2}{\|\hparam_t\|^2} =\lambda_{1}(V_{t}) $, it holds that $\lim_{t\to \infty} \hat{\alpha}_t = 1$.	
\end{proposition}

\proof{}
	This corresponds to the case where $m, M \to \lambda_1(V_{t})$ in Lemma~\ref{lem:Vinv_norm}.
	Without loss of generality, we let $M>\lambda_2(V_{t})$. Then
	$
	\hat{\alpha}_t  =(\lambda_{1}^{-1}(V_{t}) + \lambda_{d}^{-1}(V_{t})   - m{\lambda_{1}^{-1}(V_{t})\lambda_{d}^{-1}(V_{t}) })/(\lambda_{k}^{-1}(V_{t}) +  \lambda_{k+1}^{-1}(V_{t})   - {M}{\lambda_{k}^{-1}(V_{t})\lambda_{k+1}^{-1}(V_{t}) } ) \to \lambda_{1}^{-1}(V_{t})/\lambda_{1}^{-1}(V_{t}) = 1
	$. 
\endproof


To empirically confirm that Case 2 is not merely theoretical, we examine the condition outlined in Proposition \ref{prop:align}. Our focus is on the ratio
$
\zeta_t\coloneqq \frac{\|\hat{\theta}_t\|_{V_t}^2/ \|\hat{\theta}_t\|^2}{\lambda_1(V_t)}
$.  Notably, $\|\hat{\theta}_t\|_{V_t}^2 $ tends to approximate $\lambda_1 \|\hat{\theta}_t\|^2$ when $\hat{\theta}_t$ is proximate to the top eigenspace of $V_t$. Consequently, this ratio serves as a proxy of the alignment of $\hat{\theta}_t$ with the top eigenspace of $V_t$. Specifically, a $\zeta_t$ value of 1 signifies that $\hat{\theta}_t$ is within the top eigenspace.

We resort to Example 1 from Section \ref{sec:simulation}, which epitomizes the general linear bandit problem in its standard form. The empirical sequence ${\zeta_t}$ is depicted in Figure~\ref{fig:proj}.

 \begin{figure}[htp]
	\centering
		\includegraphics[width=.7\textwidth]{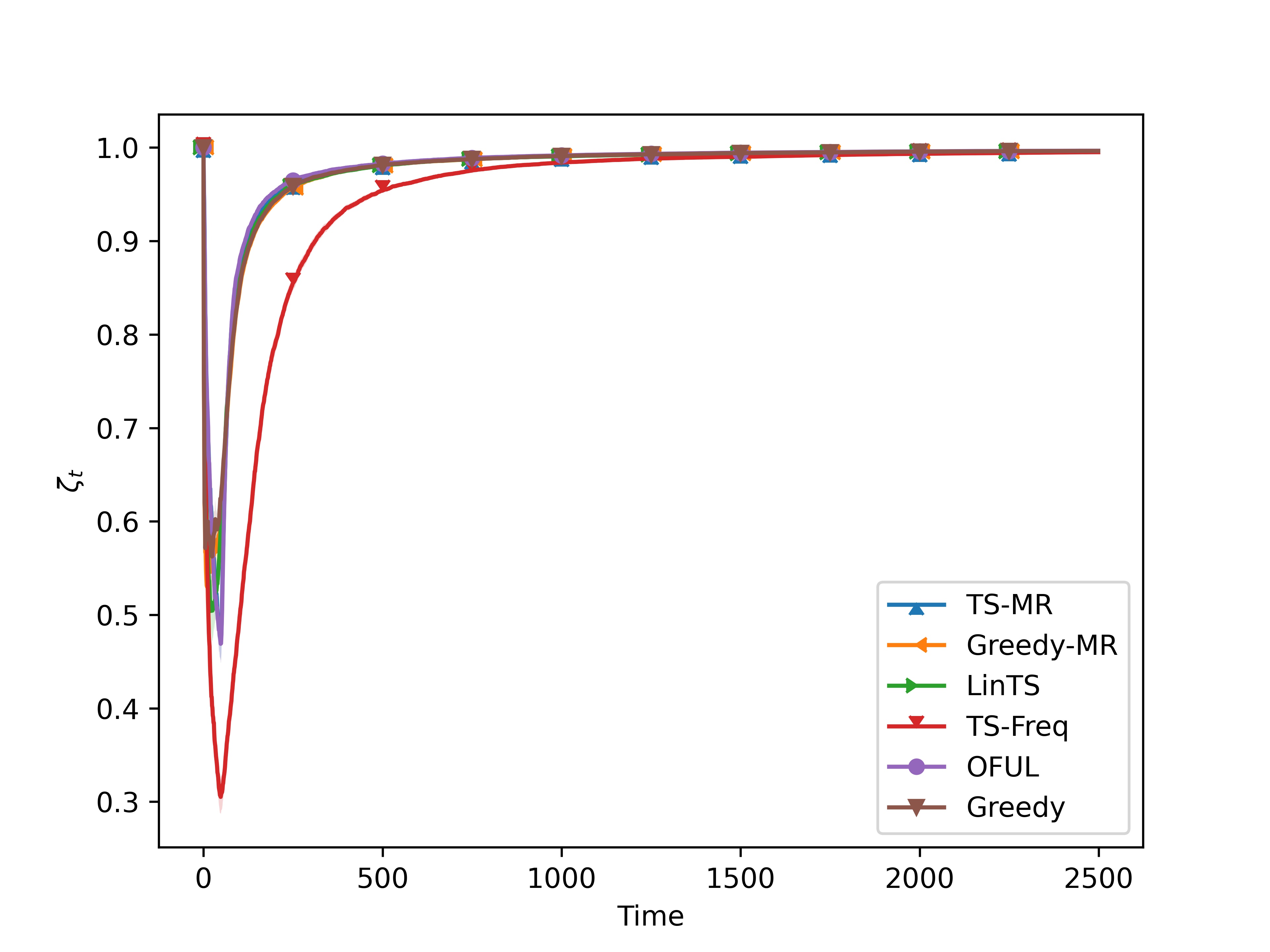} 
	\caption{Evolution of the alignment proxy $\zeta_t$ in Example 1.}
	\label{fig:proj}
\end{figure}


Figure~\ref{fig:proj} reveals that for every bandit algorithm, the value of $\zeta_t$ is notably high (exceeding 0.9) at the early stages of implementation and gradually converges towards 1. This observation validates the behavior outlined in Case 2. Furthermore, the consistency of this phenomenon across all examined bandit algorithms suggests that such a tendency is likely a characteristic of the online nature of these algorithms.

It is important to note that the initial sharp decline observed is attributed to the behavior of the regularized least squares estimator in over-parameterized scenarios, when $t$ is less than the dimension $d$.

\newpage
\section{TS-MR and greedy-MR Algorithms}
\label{sec:TS-MR}
\begin{algorithm}[!htp]
	\caption{TS-MR (Greedy-MR)}\label{alg:TS-MR}
	\begin{algorithmic}
		\REQUIRE $T$, $\delta$,  $\{\iota_t\}_{t\in [T]}$, $\mu$	
		\STATE Initialize $ V_{0} \leftarrow \lambda I_d$,  $\hat{\theta}_{1} \leftarrow 0$, $\delta^\prime \leftarrow \delta/2T$
		\FOR{t = 1, 2 , \dots, T }
		\STATE Calculate $\halpha_t$ using Theorem~\ref{thm:alpha}
		\STATE $\hat{\mu_t} \leftarrow  \hat{\alpha}_t(1+\iota_t )+1+\iota_t$ 
		\IF{$\hat{\mu}_t \leq \mu$}
		\STATE Sample $\eta_t\sim \cD^{SA}(\delta^\prime)$ (defined in Section~\ref{sec:POFUL})
		\STATE $\tilde{\theta}_{t}  \leftarrow \hat{\theta}_t +\iota_t \beta_{t,\delta^\prime, \lambdareg}^{RLS} V_t^{-\frac{1}{2}} \eta_t$
		\STATE ${x}_{t} \leftarrow \arg \max _{x \in \mathcal{X}_t}\dotp{ x}{ \tilde{\theta}_{t}}$
		\ELSE
		\STATE   ${x}_{t} \leftarrow \arg \max _{x \in \mathcal{X}_t} \arg \max _{\theta \in \cE_{t,\delta^\prime, \lambdareg}^{RLS} }\dotp{ x}{ \tilde{\theta}_{t}}$
		\ENDIF
		\STATE Observe reward $r_{t}$
		\STATE $V_{t+1} \leftarrow V_{t}+{x}_{t}  {x}_{t}^{\top}$
		\STATE $\hat{\theta}_{t+1} \leftarrow V_{t+1}^{-1}\sum_{s=1}^{t} x_s r_{s}$.
		\ENDFOR
	\end{algorithmic}
\end{algorithm}

\section{Influence of $\mu$}
\label{sec:tune_mu}
In this section, we investigate the influence of $\mu$ and discuss strategies for selecting its value. We vary the value of $\mu$ and conduct simulations on Example 2 and the Segment dataset as done in Section~\ref{sec:simulation}. The former represents a scenario where course-correction is necessary (otherwise LinTS fails), while the latter represents a practical scenario where course-correction is unnecessary. The results are shown in Figures \ref{fig:oful_frac_ex2} and \ref{fig:oful_frac_emp3}. 

 \begin{figure}[!htp]
	\centering
	\begin{subfigure}[b]{0.32\textwidth}
		\includegraphics[width=\textwidth]{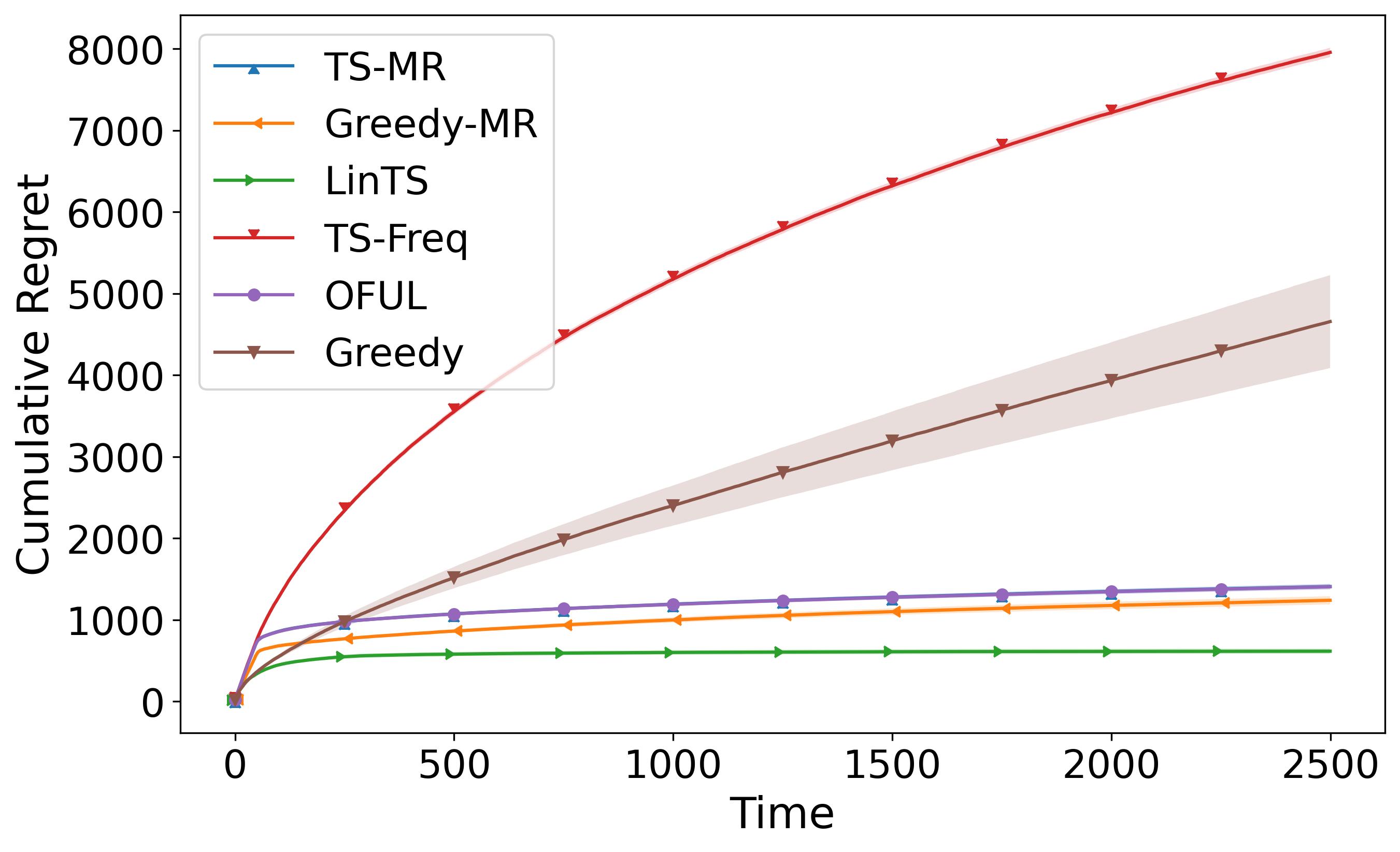} 
	\caption{example 2, $\mu = 4$}
		\label{fig:cumuregrets_ex2_mu_4}
\end{subfigure}
	\begin{subfigure}[b]{0.32\textwidth}
	\includegraphics[width=\textwidth]{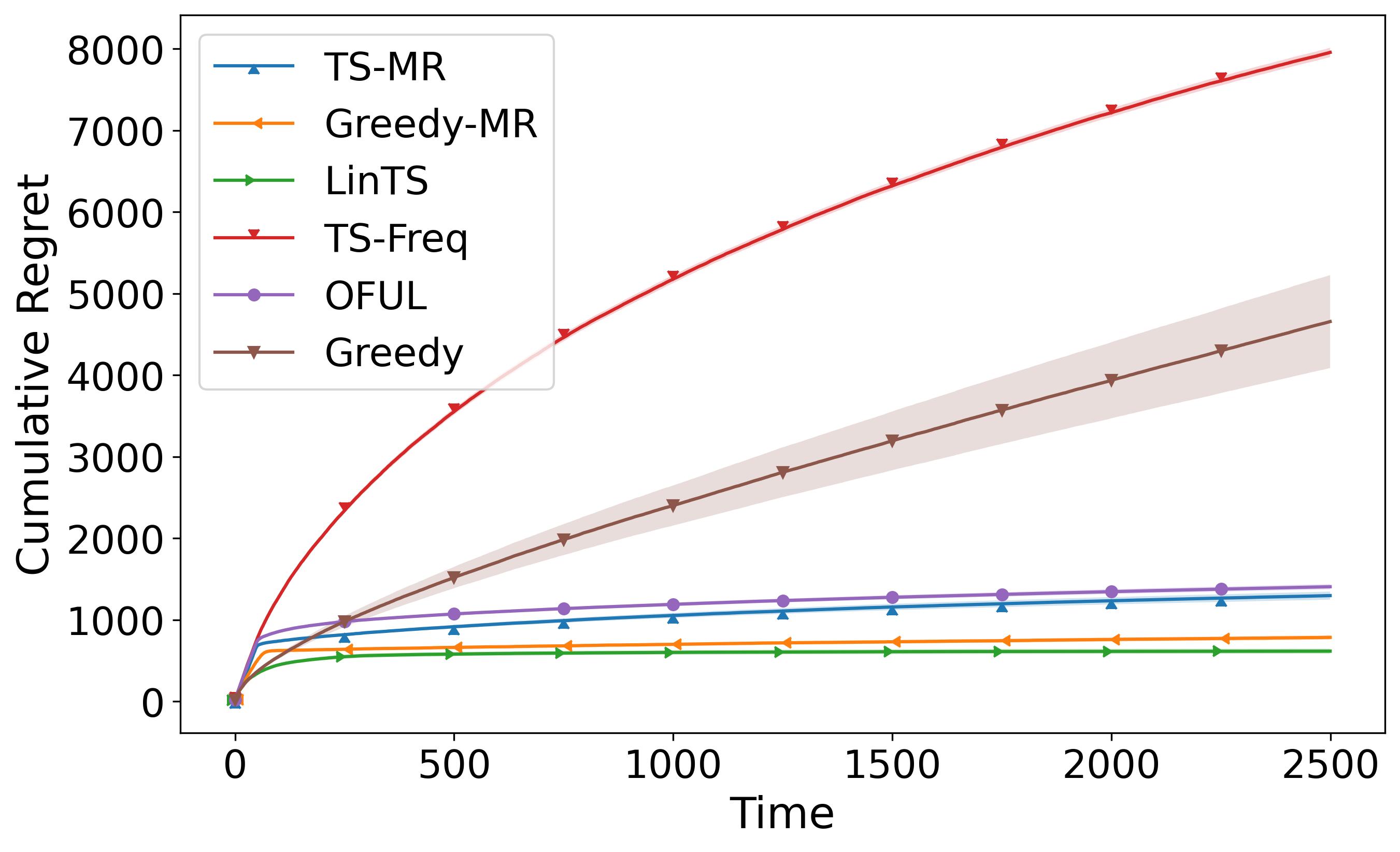} 
	\caption{example 2, $\mu = 8$}
		\label{fig:cumuregrets_ex2_mu_8}
	\end{subfigure}
 	\begin{subfigure}[b]{0.32\textwidth}
	\includegraphics[width=\textwidth]{figures/cumuregrets_ex2.jpg} 
	\caption{example 2, $\mu = 12$}
		\label{fig:cumuregrets_ex2_mu_12}
	\end{subfigure}\\
	\begin{subfigure}[b]{0.32\textwidth}
		\includegraphics[width=\textwidth]{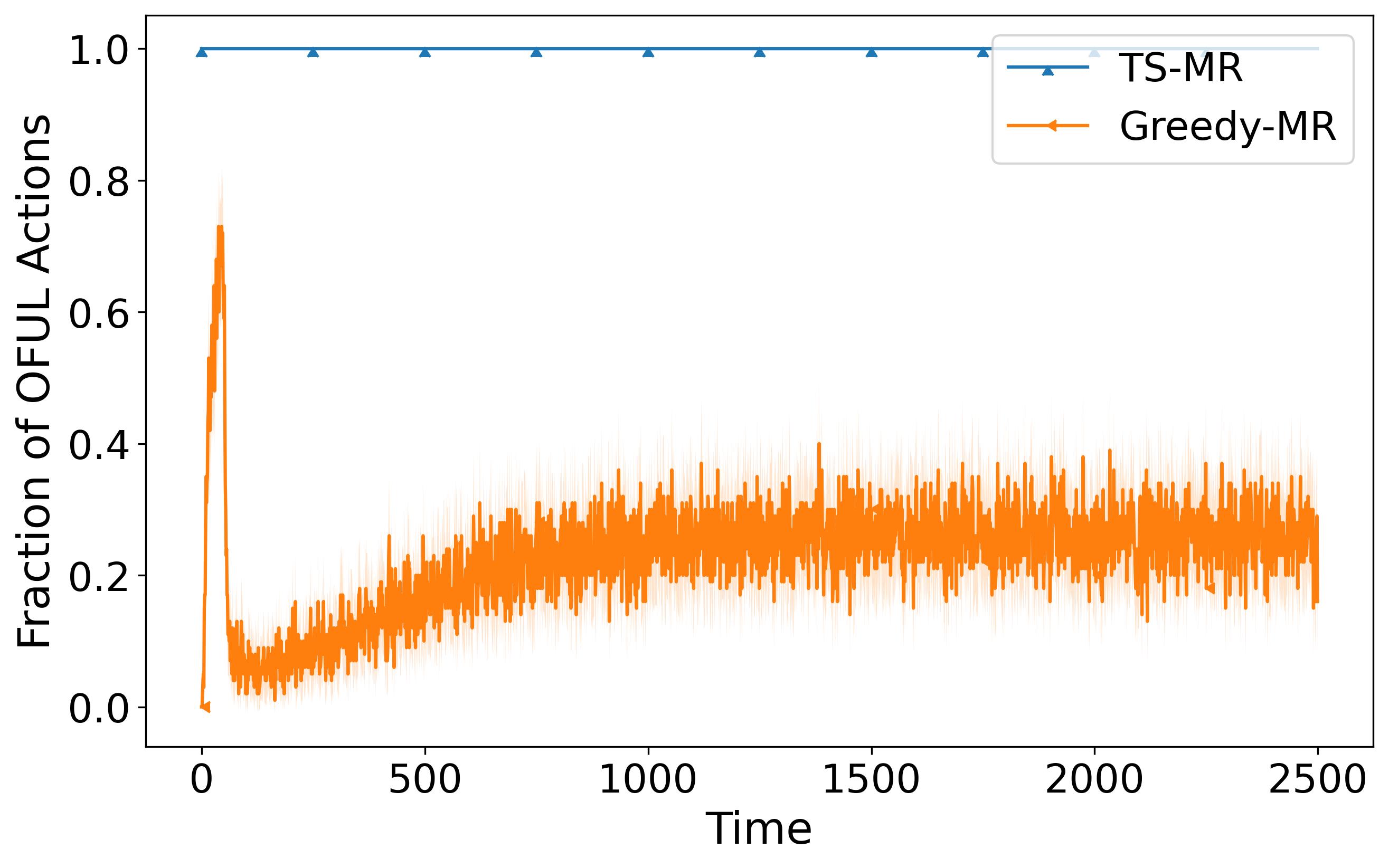} 
	\caption{example 2, $\mu = 4$}
		\label{fig:oful_frac_ex2_mu_4}
\end{subfigure}
	\begin{subfigure}[b]{0.32\textwidth}
	\includegraphics[width=\textwidth]{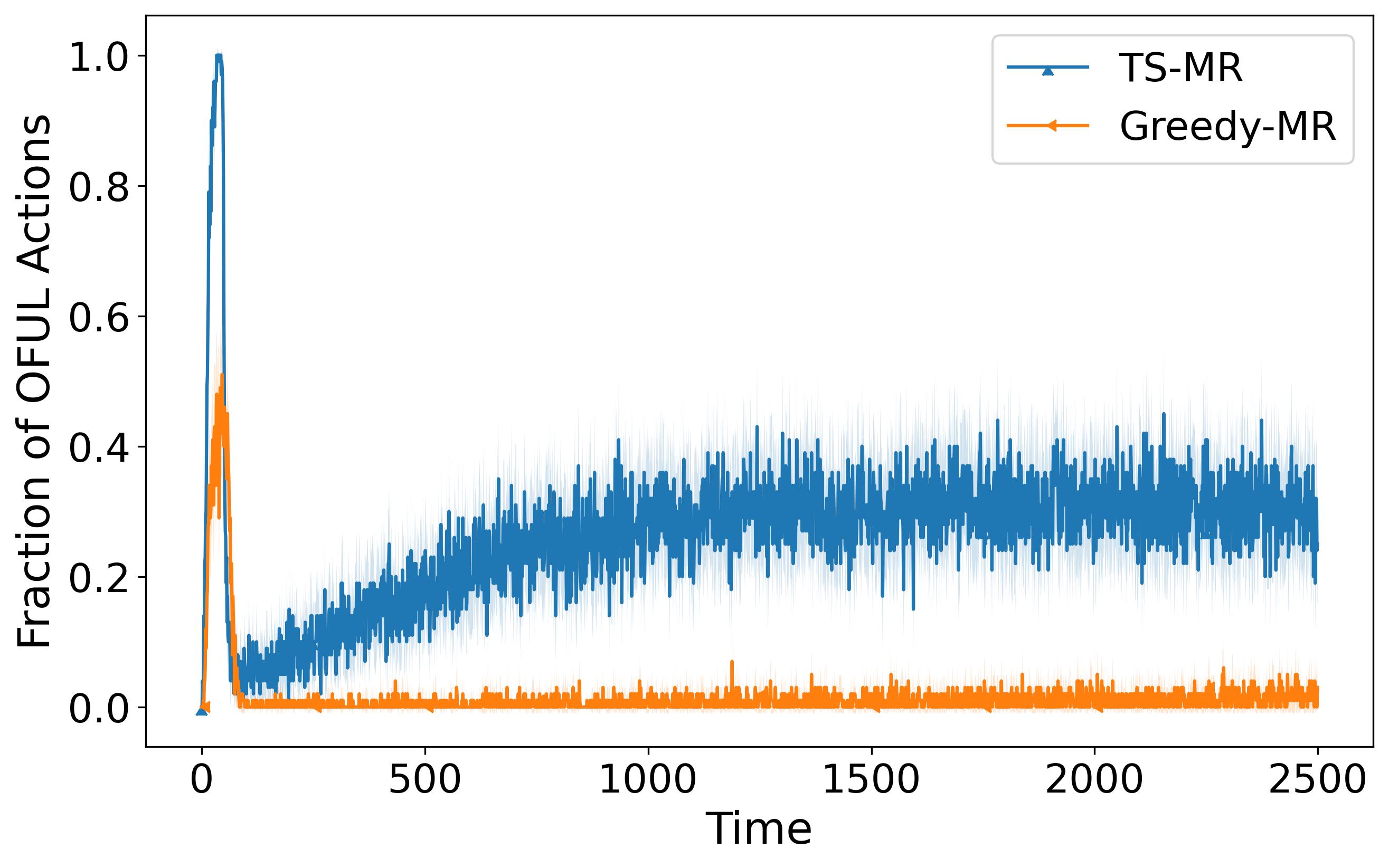} 
	\caption{example 2, $\mu = 8$}
		\label{fig:oful_frac_ex2_mu_8}
	\end{subfigure}
 	\begin{subfigure}[b]{0.32\textwidth}
	\includegraphics[width=\textwidth]{figures/oful_fraction_ex2.jpg} 
	\caption{example 2, $\mu = 12$}
		\label{fig:oful_frac_ex2_mu_12}
	\end{subfigure}
	\caption{Cumulative regret and fraction of OFUL actions of TS-MR and Greedy-MR on example 2. Shaded regions show the $\pm 2$ SE of the mean regret.}
	\label{fig:oful_frac_ex2}
\end{figure}

 \begin{figure}[htp]
	\centering
	\begin{subfigure}[b]{0.32\textwidth}
		\includegraphics[width=\textwidth]{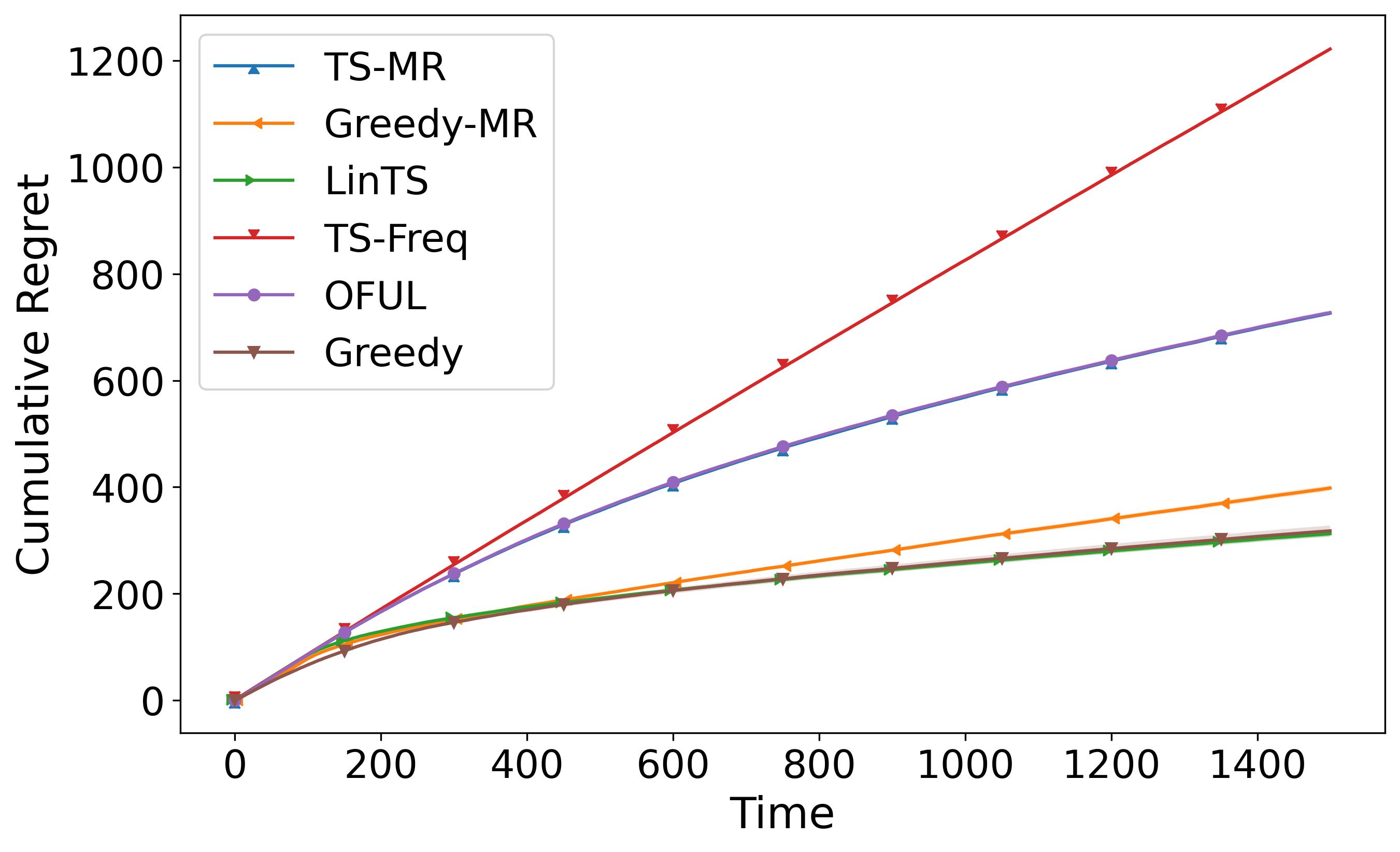} 
	\caption{Segment dataset, $\mu = 4$}
		\label{fig:cumuregrets_emp3_mu_4}
\end{subfigure}
	\begin{subfigure}[b]{0.32\textwidth}
	\includegraphics[width=\textwidth]{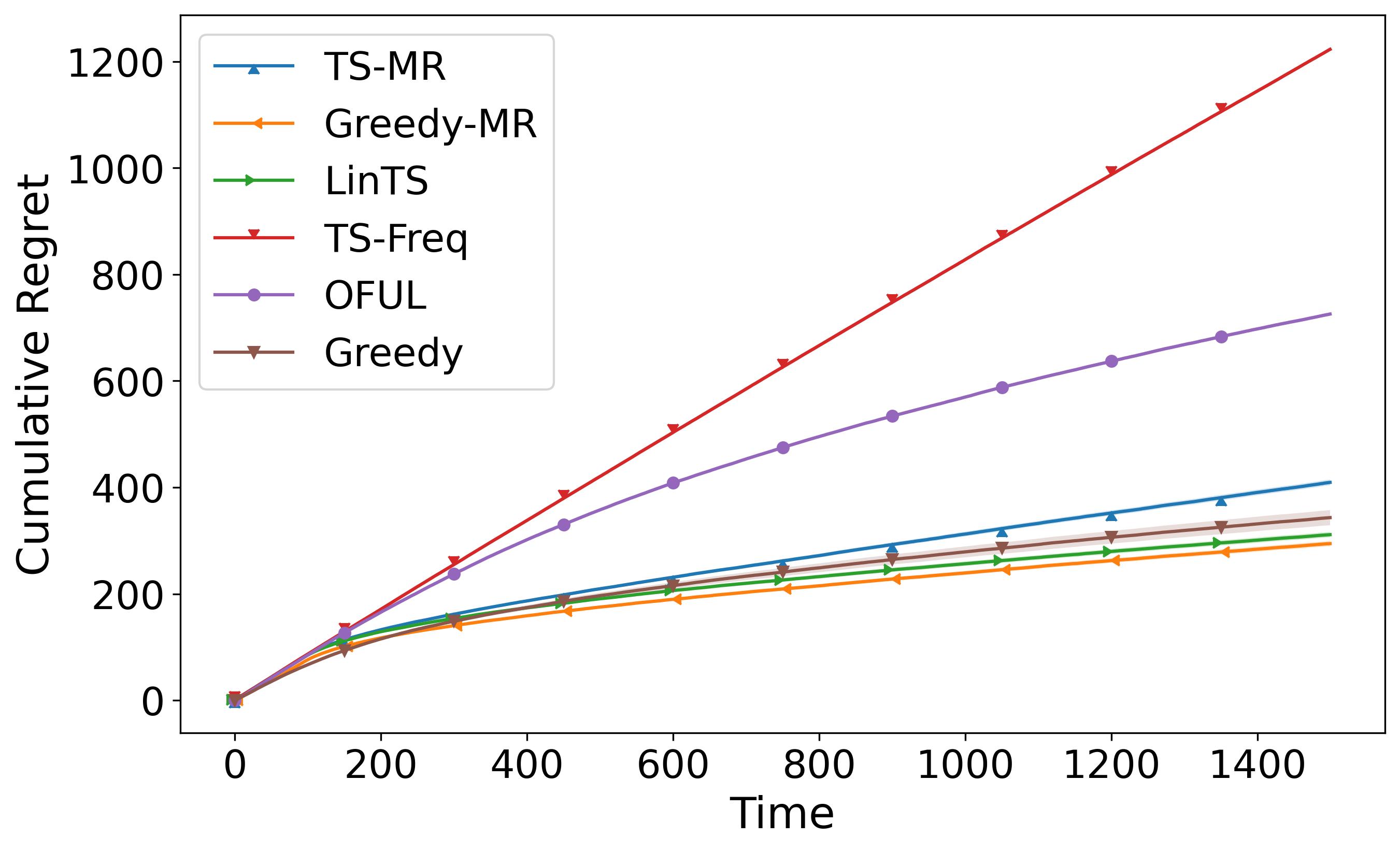} 
	\caption{Segment dataset, $\mu = 8$}
		\label{fig:cumuregrets_emp3_mu_8}
	\end{subfigure}
 	\begin{subfigure}[b]{0.32\textwidth}
	\includegraphics[width=\textwidth]{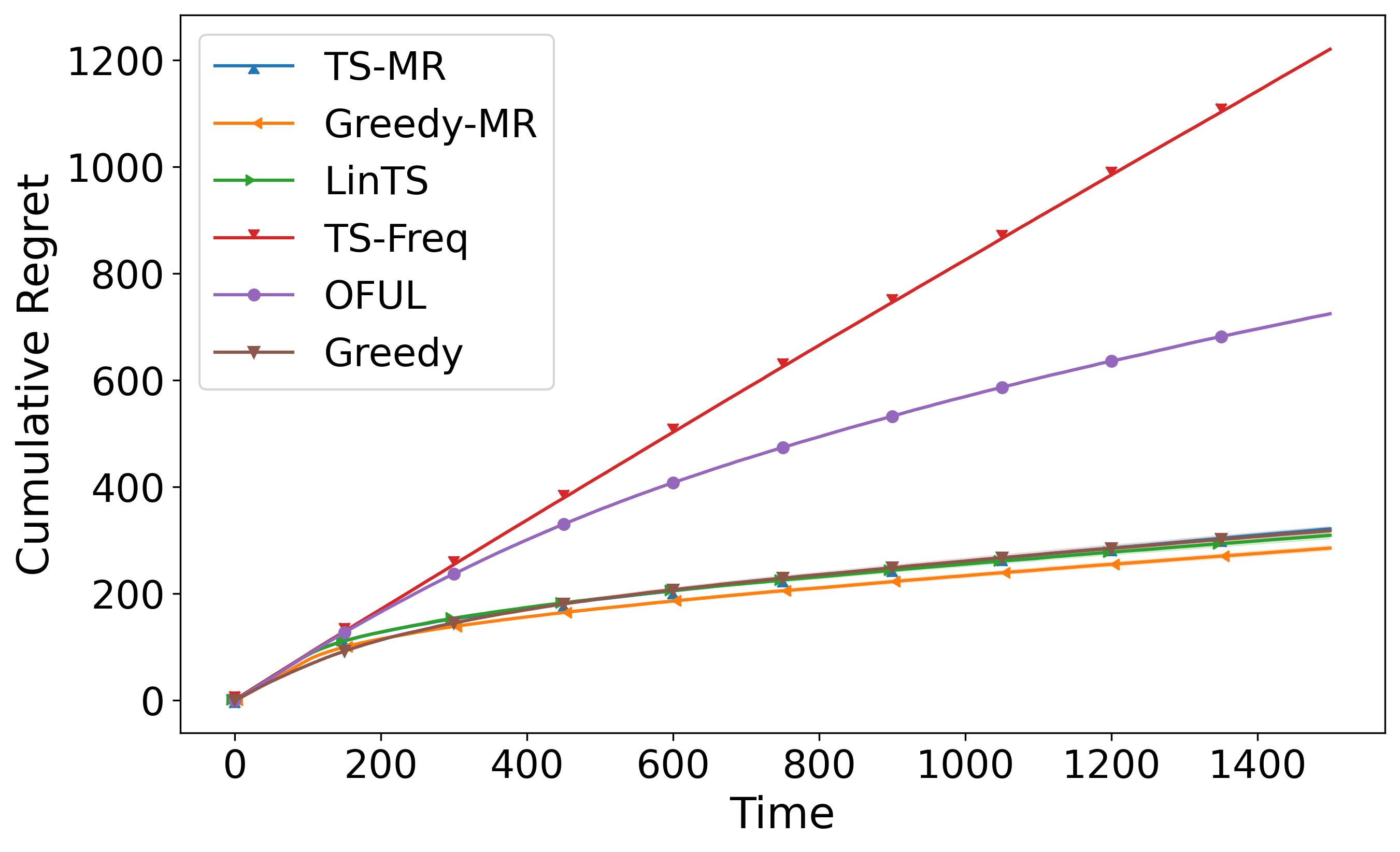} 
	\caption{Segment dataset, $\mu = 12$}
		\label{fig:cumuregrets_emp3_mu_12}
	\end{subfigure}\\
	\begin{subfigure}[b]{0.32\textwidth}
		\includegraphics[width=\textwidth]{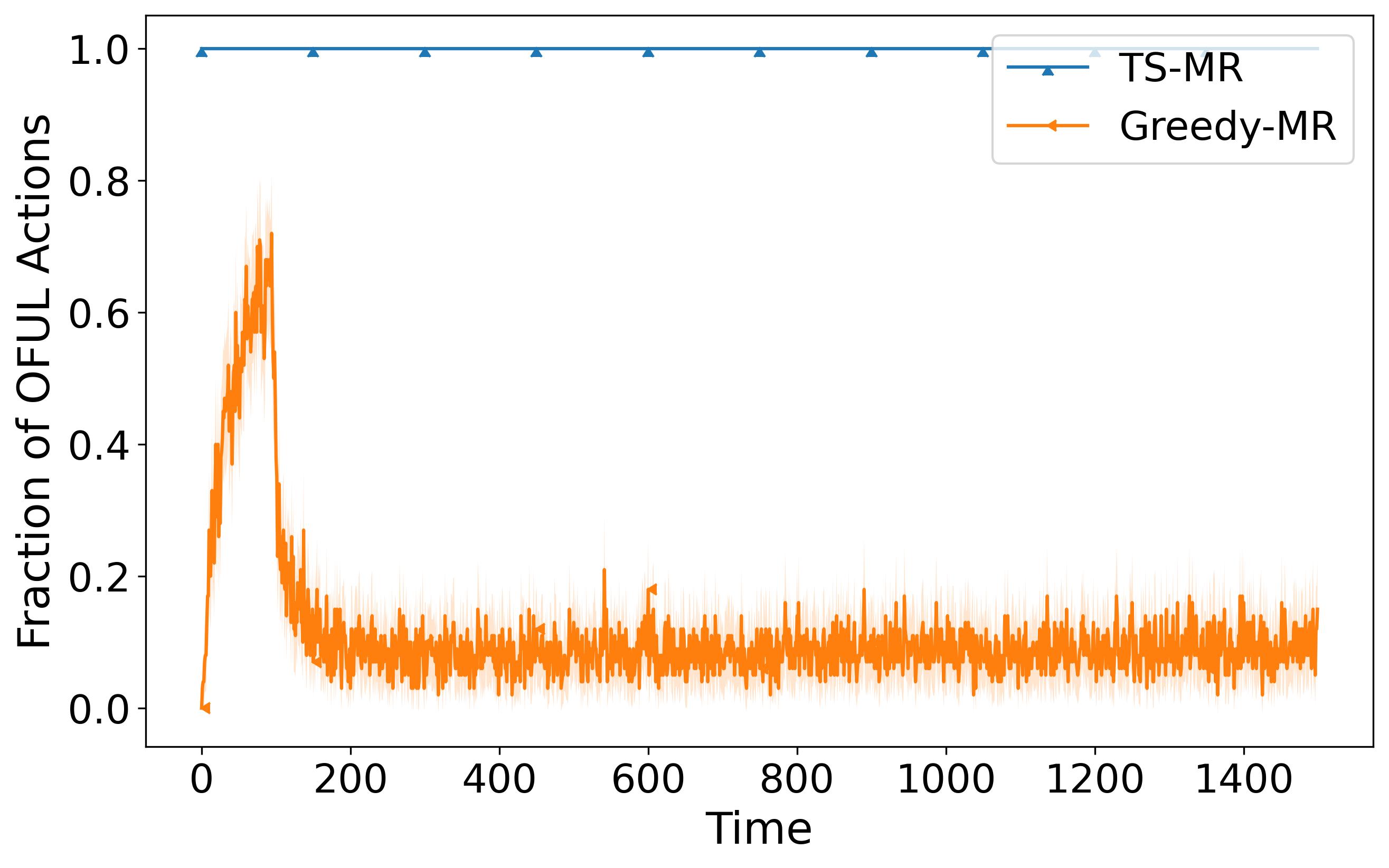} 
	\caption{Segment dataset, $\mu = 4$}
		\label{fig:oful_frac_emp3_mu_4}
\end{subfigure}
	\begin{subfigure}[b]{0.32\textwidth}
	\includegraphics[width=\textwidth]{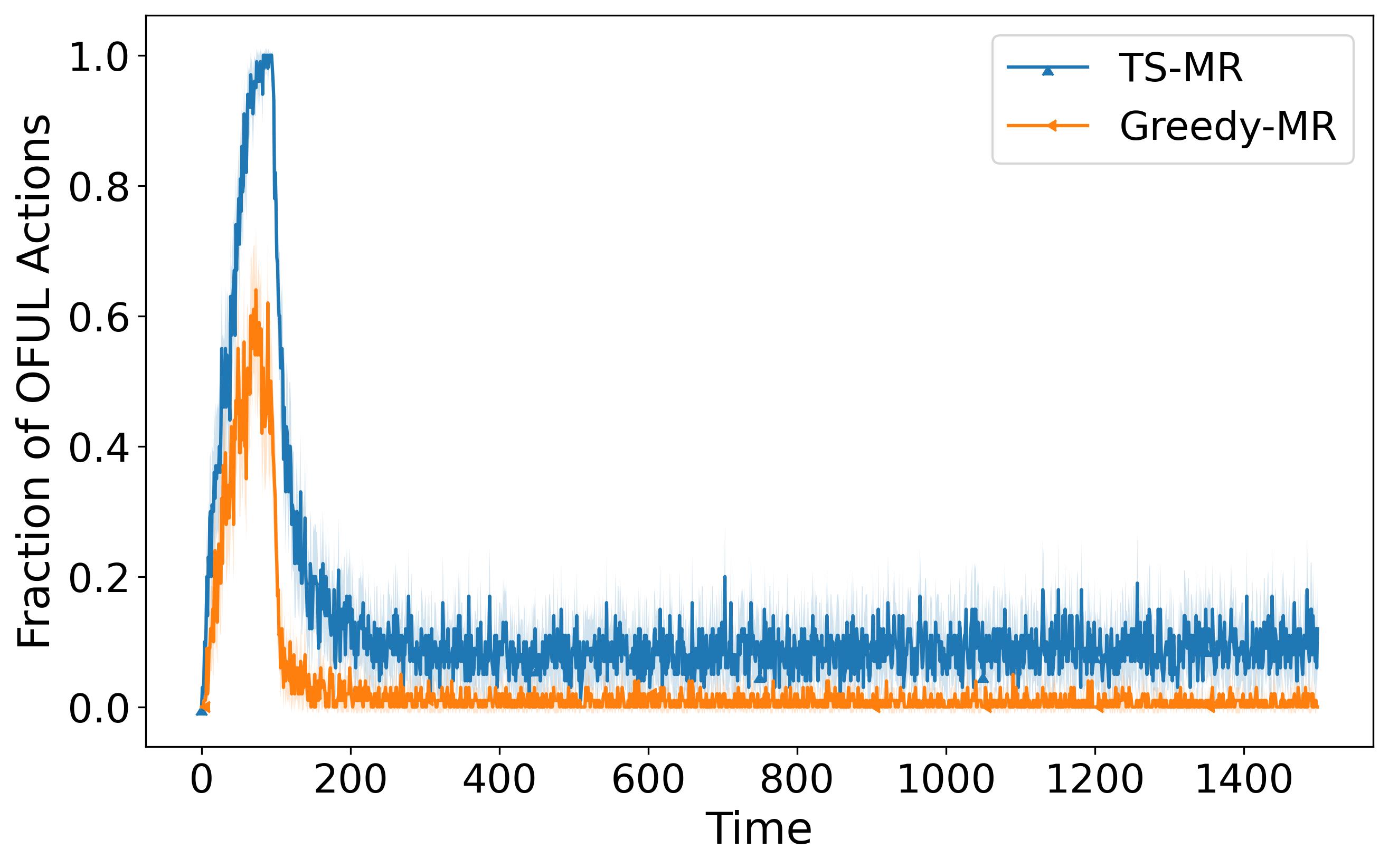} 
	\caption{Segment dataset, $\mu = 8$}
		\label{fig:oful_frac_emp3_mu_8}
	\end{subfigure}
 	\begin{subfigure}[b]{0.32\textwidth}
	\includegraphics[width=\textwidth]{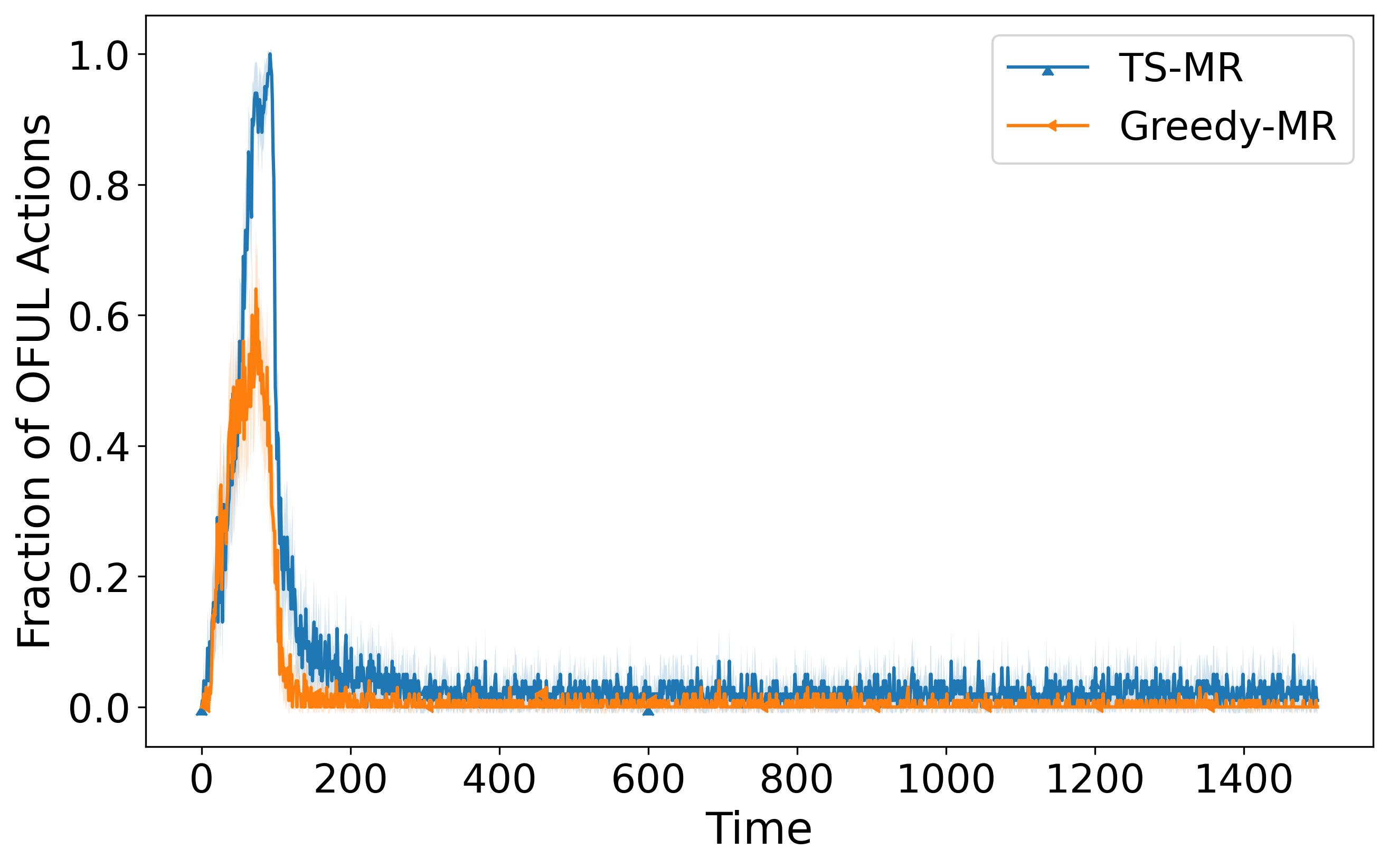} 
	\caption{Segment dataset, $\mu = 12$}
		\label{fig:oful_frac_emp3_mu_12}
	\end{subfigure}
	\caption{Cumulative regret and fraction of OFUL actions of TS-MR and Greedy-MR on Segment datasets. Shaded regions show the $\pm 2$ SE of the mean regret.}
	\label{fig:oful_frac_emp3}
\end{figure}

We have the following observations on the influence of $\mu$:
\begin{itemize}
    \item In general, as $\mu$ increases, the fraction of OFUL actions decreases. For a sufficiently small $\mu$, the algorithms become OFUL (e.g., TS-MR in Figures~\ref{fig:cumuregrets_ex2_mu_4} and \ref{fig:cumuregrets_emp3_mu_4}). For a large $\mu$, the algorithms are similar to the original algorithms (e.g., TS-MR and Greedy-MR in Figures~\ref{fig:cumuregrets_ex2_mu_12} and \ref{fig:cumuregrets_emp3_mu_12}). 
    \item By optimizing the choice of $\mu$, the performance of the corrected algorithm might outperform both OFUL and the base algorithm (e.g., Greedy-MR in Figures~\ref{fig:cumuregrets_ex2_mu_8} and \ref{fig:cumuregrets_emp3_mu_8}). This demonstrates that TS/Greedy-MR are not merely naive interpolations between TS/Greedy and OFUL, and that proper exploration in the initial stage benefits the long-term performance of the algorithms.
    
    \item As long as $\mu$ isn't too small,  the performance of the course-corrected algorithm are relatively robust to the choice of $\mu$. The simulations show $\mu\in[8,12]$ is a good initial choice.
\end{itemize}

\paragraph{Empirical guidelines and heuristics for choosing $\mu$.} In practice, we recommend setting $\mu$ to a moderate value, typically within the range $[8,12]$, as validated by our simulation results. The selection of $\mu$ should be guided by heuristics that ensure an appropriate fraction of course-corrected exploration during the initial stage. A suitable $\mu$ value is indicated when the fraction of OFUL actions is high at the beginning and gradually decreases to and maintains a low level. If the fraction of OFUL actions remains consistently high, increasing $\mu$ can help reduce computational costs. Conversely, if very few OFUL actions are executed during the initial stage, decreasing $\mu$ may be beneficial. Finally, we note that since algorithmic performance remains robust across moderate values of $\mu$, the precise selection of $\mu$ is unlikely to be a significant practical concern.

\end{APPENDICES}

\bibliographystyle{ormsv080}
\bibliography{asymptotic-bandit}

\end{document}